\documentclass{article} 

\usepackage{amsfonts}
\usepackage{amssymb}
\usepackage{amsmath}
\usepackage{amsthm}
\usepackage{latexsym}
\usepackage{verbatim}
\usepackage{epsfig}
\usepackage{color}
										
\usepackage{graphicx}
\usepackage{amsbsy}
\usepackage{amscd}
\usepackage{bm}

\newcommand{\R}{{\mathbb R}}

\DeclareMathOperator*{\argmin}{argmin}

\DeclareMathOperator*{\p}{\mathbf{P}}

\providecommand{\wh}[1]{\widehat{#1}}

\providecommand{\nnorm}[1]{ \lVert#1 \rVert}

\providecommand{\mc}[1]{\mathcal#1}
\providecommand{\T}{\top}

\newcommand{\aff}{\text{aff}}

\newcommand{\cls}[1]{_{:,#1}}

\newcommand{\blanco}[1]{  }

\newcommand{\deriv}[3]{%
\ifthenelse{#1 = 1}{\frac{d\,#2}{d\,#3}}{\frac{d^{{#1}} #2}{d{#3}^{{#1}}}}
}

\newcommand{\partials}[3]{%
\ifthenelse{#1 = 1}{\frac{\partial\,#2}{\partial\,#3}}{\frac{\partial^{#1}
    #2}{\partial#3^{#1}}}
}

\newtheorem{theo}{Theorem}
\newtheorem{propo}{Theorem}

\newtheorem{theoA}{Theorem}[section]

\newtheorem{corollary}{Theorem}

\newtheorem{theoApp}[theoA]{Theorem}

  \newtheorem{prop}[propo]{Proposition}
  \newtheorem{corro}[corollary]{Corollary}

\newenvironment{bew}{\begin{proof}[Proof]}{\end{proof}}

\usepackage{nips13submit_e,times}
\usepackage{algorithm}
\usepackage{algorithmic}
\usepackage{url}

\title{Matrix factorization with Binary Components}

\author{Martin Slawski, Matthias Hein and Pavlo Lutsik\\ 
Saarland University\\
\texttt{\{ms,hein\}@cs.uni-saarland.de}, \texttt{p.lutsik@mx.uni-saarland.de}  
}

\nipsfinalcopy 

\begin{document}

\maketitle

\begin{abstract}
Motivated by an application in computational biology, we consider 
low-rank matrix factorization with $\{0,1\}$-constraints on one
of the factors and optionally convex constraints on the second one. In
addition to the non-convexity shared with other
matrix factorization schemes, our problem is further complicated by a
combinatorial constraint set of size $2^{m \cdot r}$, where $m$ is the
dimension of the data points and $r$ the rank of the factorization. Despite
apparent intractability, we provide $-$ in the line of recent work on
non-negative matrix factorization by Arora et al.~(2012)$-$ an algorithm that
provably recovers the underlying factorization in the exact case with {\small$O(m r 2^r + mnr + r^2 n)$}
operations for $n$ datapoints. To obtain this result, we use theory
around the Littlewood-Offord lemma from combinatorics.    
\end{abstract}

\section{Introduction}\label{sec:intro}

Low-rank matrix factorization techniques like the singular value decomposition
(SVD) constitute an important tool in data analysis yielding a compact
representation of data points as linear combinations of a comparatively small
number of 'basis elements' commonly referred to as \emph{factors}, \emph{components} or
\emph{latent variables}. Depending on the specific application, the basis elements
may be required to fulfill additional properties, e.g.~non-negativity 
\cite{Paatero1994, Lee99}, smoothness \cite{Ram2006} or sparsity
\cite{Bach2008, Witten2009}. In the present paper, we consider the
case in which the basis elements are constrained to be binary, i.e.~we aim at
factorizing a real-valued data matrix $D$ into a product $T A$ with $T \in
\{0,1\}^{m \times r}$ and $A \in \R^{r \times n}$, $r \ll \min\{m,n\}$. Such decomposition arises
e.g.~in blind source separation in wireless communication with binary source
signals \cite{Veen1997}; in network inference from gene expression data
\cite{Liao2003, Tu2012}, where $T$ encodes connectivity of
transcription factors and genes; in unmixing of cell mixtures from DNA
methylation signatures 
\cite{Houseman2012short} in which case $T$ represents presence/absence of
methylation; or in clustering with overlapping clusters with $T$ as a matrix
of 
cluster assignments \cite{Banerjee2005, Segal2003}.  
\\
Several other matrix factorizations involving binary matrices
have been proposed in the literature. In \cite{Schein2003} and \cite{Kaban2008} matrix factorization
for binary input data, but non-binary factors $T$ and $A$ is discussed, whereas
a factorization $T W A$ with both $T$ and $A$ binary and
real-valued $W$ is proposed in \cite{Meeds2007}, which is more restrictive 
than the model of the present paper. The model in \cite{Meeds2007} in turn encompasses binary matrix factorization as
proposed in \cite{Zhang2007}, where all of $D$, $T$ and $A$ are constrained to
be binary. It is important to note that this ine of research is
fundamentally different from Boolean matrix factorization \cite{Miettinen2006}, which is sometimes also referred to as
binary matrix factorization.\\
A major drawback of matrix factorization schemes is non-convexity. As a
result, there is in general no algorithm that is guaranteed to compute the
desired factorization. Algorithms such as block coordinate descent, EM, MCMC, etc.~commonly employed in practice lack theoretical guarantees
beyond convergence to a local minimum. Substantial progress in this regard has been achieved
recently for non-negative matrix factorization (NMF) by Arora et al.~\cite{Arora2012} and
follow-up work in \cite{Bittdorf2012}, where it is shown that under certain
additional conditions, the NMF problem can be solved globally optimal by means of linear
programming. Apart from being a non-convex problem, the matrix factorization studied in
the present paper is further complicated by the
$\{0,1\}$-constraints imposed on the left factor $T$, which yields a
combinatorial optimization problem that appears to be computationally
intractable except for tiny dimensions $m$ and $r$ even in case the right factor $A$ were already
known. Despite the obvious hardness of the problem, we present as our main
contribution an algorithm that provably provides an exact factorization $D =
TA$ whenever such factorization exists. Our algorithm has exponential
complexity only in the rank $r$ of the factorization, but scales linearly in
$m$ and $n$. In particular, the problem remains tractable even for large
values of $m$ as long as $r$ remains small. We extend the algorithm to the approximate case 
$D \approx T A$ and empirically show superior performance relative to heuristic approaches to the problem. Moreover, we
establish uniqueness of the exact factorization under the separability
condition from the NMF literature \cite{Arora2012, DonohoStodden03}, or alternatively with high probability for $T$ drawn
uniformly at random. As a corollary, we obtain that at least for these two
models, the suggested algorithm continues to be fully applicable if additional
constraints e.g.~non-negativity, are imposed on the right factor $A$. We
demonstrate the practical usefulness of our approach in
unmixing DNA methylation signatures of blood samples \cite{Houseman2012short}.\\
\textbf{Notation.} For a matrix $M$ and index sets $I,J$, $M_{I,J}$
denotes the submatrix corresponding to $I$ and $J$; $M_{I,:}$ and $M_{:,J}$ denote the
submatrices formed by the rows in $I$ respectively columns in $J$. We write
$[M;M']$ and $[M, M']$ for the row- respectively column-wise concatenation
of $M$ and $M'$. The affine hull generated by the
columns of $M$ is denoted by $\aff(M)$. The symbols $\bm{1}$/$0$ denote vectors or matrices of ones/zeroes and
$I$ denotes the identity matrix. We use $|\cdot|$ for the cardinality of a
set.\\
\textbf{Appendix.} The appendix contains all proofs, additional comments
and experimental results.    

\section{Exact case}\label{gen_inst}

We start by considering the exact case, i.e.~we suppose that a
factorization having the desired properties exists. We first discuss the geometric ideas underlying our basic approach for recovering such
factorization from the data matrix before presenting conditions under which
the factorization is unique. It is shown that the question of uniqueness as
well as the computational performance of our approach is intimately connected to the Littlewood-Offord problem in
combinatorics \cite{Erdos1951}.\\
\\ 
\textbf{2.1 Problem formulation.} Given $D \in \R^{m \times n}$, we consider the following problem.
\begin{equation}\label{eq:exactfactorize}
\text{find} \; T \in \{0,1\}^{m \times r} \; \, \text{and} \; \, A \in \R^{r
  \times n}, \, A^{\T} \bm{1}_r = \bm{1}_n \; \; \text{such that} \; \, D = TA.
\end{equation}
The columns $\{ T\cls{k} \}_{k = 1}^r$ of $T$, which are vertices of the
hypercube $[0,1]^m$, are referred to as components. 
The requirement $A^{\T} \bm{1}_r = \bm{1}_n$ entails that the columns of $D$ are affine instead of linear combinations of
the columns of $T$. This additional constraint is not essential to our
approach; it is imposed for reasons of presentation, in order to avoid
that the origin is treated differently from the other vertices of $[0,1]^m$,
because otherwise the zero vector could be dropped from $T$, leaving
the factorization unchanged. We further assume w.l.o.g.~that $r$ is minimal,
i.e.~there is no factorization of the form \eqref{eq:exactfactorize}
with $r' < r$, and in turn that the columns of $T$ are affinely independent, i.e.~$\forall
\lambda \in \R^r, \, \lambda^{\T} \bm{1}_r = 0$, $T \lambda = 0$ implies that
$\lambda = 0$. Moreover, it is assumed that $\text{rank}(A) = r$. This ensures the
existence of a submatrix $A\cls{\mc{C}}$ of $r$ linearly independent columns
and of a corresponding submatrix of $D\cls{\mc{C}}$ of affinely independent
columns, when combined with the affine independence of the columns of $T$:
\begin{equation}\label{eq:affineindependenceD}
\forall \lambda \in \R^r, \, \lambda^{\T} \bm{1}_r = 0:\, D\cls{\mc{C}} \lambda = 0  \, \Longleftrightarrow \; T (A\cls{\mc{C}} \lambda) = 0 \;
\Longrightarrow A\cls{\mc{C}} \lambda = 0 \Longrightarrow \lambda = 0, 
\end{equation}
using at the second step that $\bm{1}_r^{\T} A\cls{\mc{C}} \lambda = \bm{1}_r^{\T} \lambda = 0$
and the affine independence of the $\{T\cls{k} \}_{k=1}^r$. Note that the
assumption $\text{rank}(A) = r$ is natural; otherwise, the data
would reside in an affine subspace of lower dimension so that $D$ would
not contain enough information to reconstruct $T$.\\
\\ 
\textbf{2.2 Approach.} 
Property \eqref{eq:affineindependenceD} already provides the entry point of our
approach. From $D = T A$, it is obvious that $\aff(T) \supseteq
\aff(D)$. Since $D$ contains the same number of affinely independent columns
as $T$, it must also hold that $\aff(D) \supseteq \aff(T)$, in particular 
$\aff(D) \supseteq \{ T\cls{k}\}_{k = 1}^r$. Consequently, \eqref{eq:exactfactorize} can in principle be solved by enumerating all
vertices of $[0,1]^m$ contained in $\aff(D)$ and selecting a maximal affinely
independent subset thereof (see Figure 1). This procedure, however, is exponential in the
dimension $m$, with $2^m$ vertices to be checked for containment in $\aff(D)$
by solving a linear system. Remarkably, the following observation
along with its proof, which prompts Algorithm \ref{alg:1} below, shows that
the number of elements to be checked can be reduced to $2^{r-1}$ irrespective
of $m$.
\begin{prop}\label{prop:basic} The affine subspace $\text{aff}(D)$ contains no more than
  $2^{r-1}$ vertices of $[0,1]^m$. Moreover, Algorithm \ref{alg:1} provides
  all vertices contained in $\text{aff}(D)$.
\end{prop}
\begin{center}
\begin{algorithm}
\caption{\textsc{FindVertices} \textsc{exact}}\label{alg:1}
\algsetup{indent=0em}
\begin{algorithmic}
\STATE \begin{enumerate}\item[1.] Fix $p \in \aff(D)$ and compute $P = [D\cls{1} -p,\ldots,D\cls{n} - p]$.
     \end{enumerate}   
\STATE \begin{enumerate}\item[2.] Determine $r-1$ linearly independent columns
  $\mc{C}$ of $P$, obtaining $P\cls{\mc{C}}$ and subsequently $r-1$ linearly independent
  rows $\mc{R}$, obtaining $P_{\mc{R},\mc{C}} \in \R^{r-1 \times r-1}$.
\end{enumerate}
\STATE \begin{enumerate}\item[3.] Form $Z = P\cls{\mc{C}}
  (P_{\mc{R},\mc{C}})^{-1} \in \R^{m \times r-1}$ and $\wh{T} = Z (B^{(r-1)}  -
  p_{\mc{R}} \bm{1}_{2^{r-1}}^{\T}) + p \bm{1}_{2^{r-1}}^{\T} \in \R^{m \times
  2^{r-1}}$, where
the columns of $B^{(r-1)}$ correspond to the elements of $\{0,1\}^{r-1}$.\end{enumerate}
\STATE \begin{enumerate}\item[4.] Set $\mc{T} = \emptyset$. For $u =
  1,\ldots,2^{r-1}$, if $\wh{T}\cls{u} \in \{0,1 \}^m$ set $\mc{T} = \mc{T}
  \cup \{ \wh{T}\cls{u} \}$.
\end{enumerate}
\STATE  \begin{enumerate}\item[5.] Return $\mc{T} = \{0,1\}^m \cap \aff(D)$. \end{enumerate}
\end{algorithmic}
\end{algorithm} 
\begin{algorithm}
\caption{\textsc{BinaryFactorization} \textsc{exact}}\label{alg:2}
\begin{algorithmic}
\STATE \begin{enumerate}\item[1.] Obtain $\mc{T}$ as output from \textsc{FindVertices Exact}($D$)
     \end{enumerate}   
\STATE \begin{enumerate}\item[2.] Select $r$ affinely independent elements of
  $\mc{T}$ to be used as columns of $T$. 
\end{enumerate}
\STATE \begin{enumerate}\item[3.] Obtain $A$ as solution of the linear system
  $[\bm{1}_r^{\T}; T] A = [\bm{1}_{n}^{\T}; D]$. 
\end{enumerate}
\STATE  \begin{enumerate}\item[4.] Return $(T,A)$ solving problem \eqref{eq:exactfactorize}. \end{enumerate}
\end{algorithmic}
\end{algorithm}
\end{center}
\begin{minipage}{0.19\textwidth}
\vspace{-0.3cm}
\begin{tabular}{ll}
\includegraphics[height = 0.1\textheight]{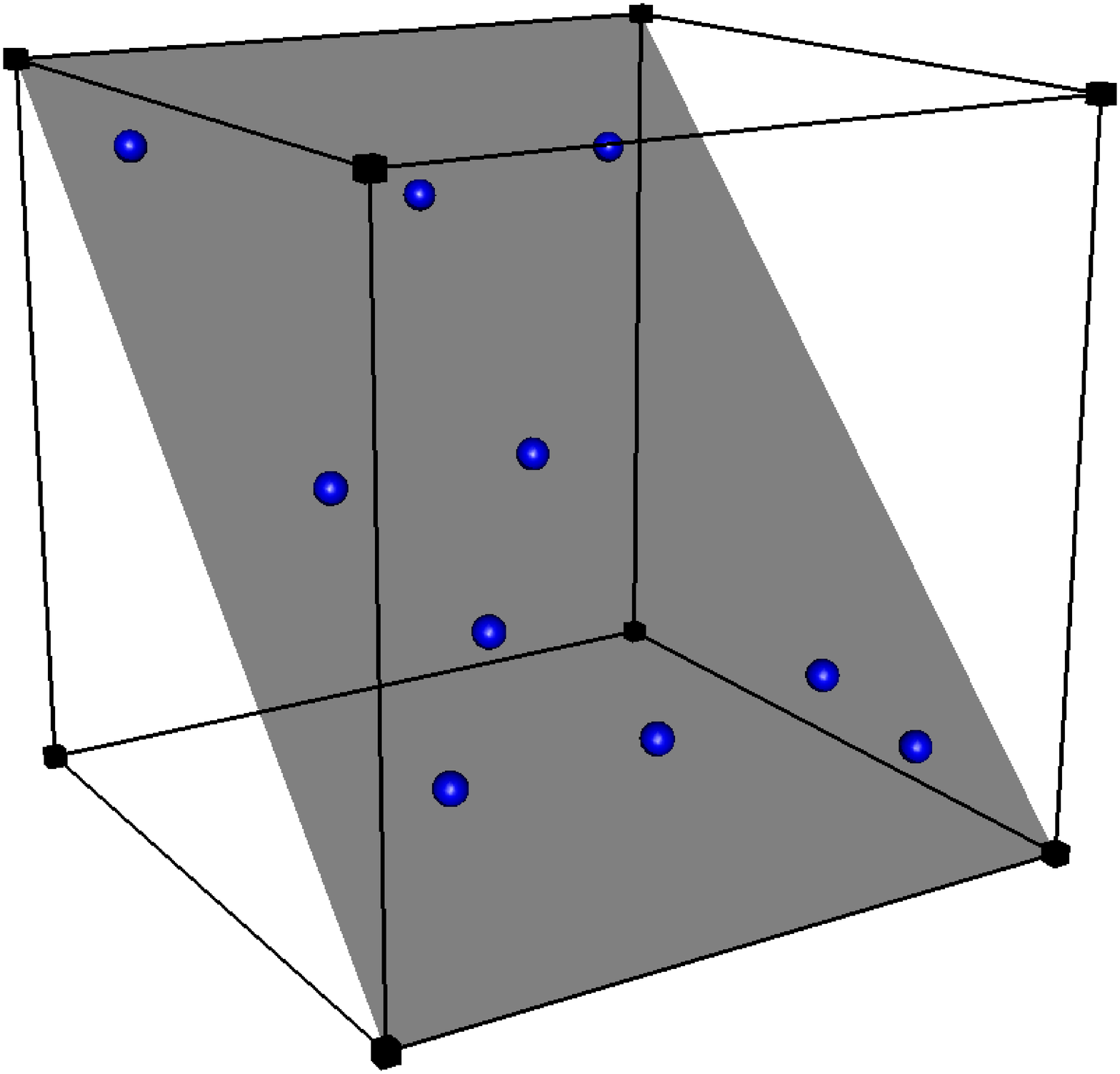}
& \includegraphics[height =0.1\textheight]{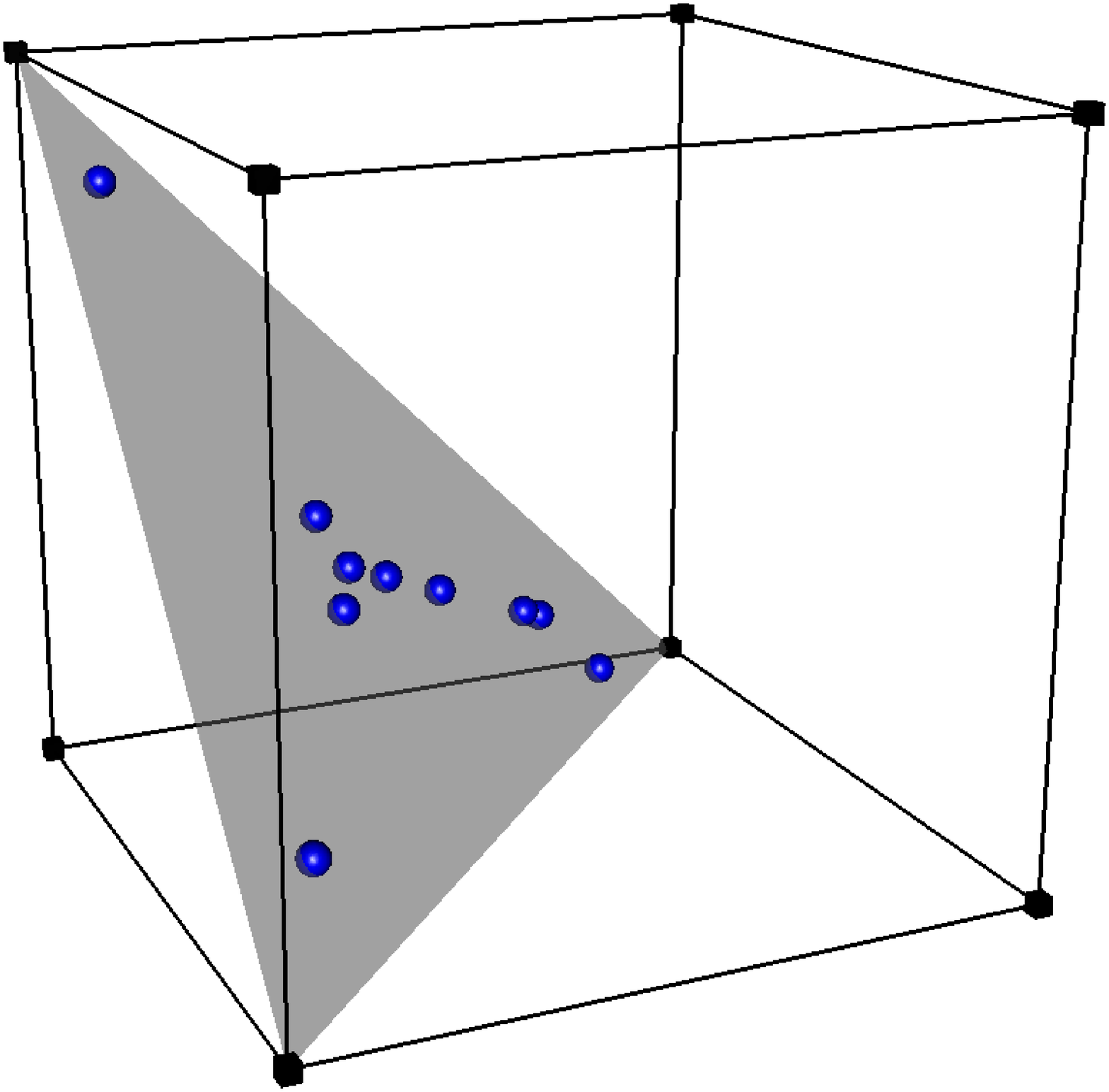}
\end{tabular}
\end{minipage}
\begin{flushright}
\begin{minipage}{0.59\textwidth}
\vspace{-0.13\textheight}
\setcounter{figure}{1}
Figure 1: Illustration of the geometry underlying our approach in dimension $m
= 3$. Dots represent data points and the shaded areas their affine hulls
$\aff(D) \cap [0,1]^m$.\\ Left: $\aff(D)$ intersects with $r + 1$ vertices of $[0,1]^m$.\\ Right: $\aff(D)$ intersects with precisely $r$ vertices.
\end{minipage}
\end{flushright}


\vspace{-0.5cm}  
\textbf{Comments.} In step 2 of Algorithm \ref{alg:1}, determining the rank
of $P$ and an associated set of linearly independent columns/rows can be done
by means of a rank-revealing QR factorization \cite{Gu1996, Golub1996}. The
crucial step is the third one, which is a compact description of first solving the linear systems $P_{\mc{R},\mc{C}}
\lambda = b - p_{\mc{R}}$ for all $b \in \{0,1\}^{r-1}$ and back-substituting the
result to compute candidate vertices $P_{:,\mc{C}} \lambda + p$ stacked into
the columns of $\wh{T}$; the addition/subtraction of $p$ is merely because
we have to deal with an affine instead of a linear subspace, in which $p$
serves as origin. In step 4, the pool of $2^{r-1}$ 'candidates' is filtered, yielding $\mc{T} = \text{aff}(D) \cap \{0,1\}^m$.\\ 
Determining $\mc{T}$ is the hardest part in solving the matrix factorization problem
\eqref{eq:exactfactorize}.  Given $\mc{T}$, the solution can be obtained after
few inexpensive standard operations. Note that step 2 in Algorithm \ref{alg:2} is not necessary if one does not aim
at finding a minimal factorization, i.e.~if it suffices to have $D = T A$ with
$T \in \{0,1\}^{m \times r'}$ but $r'$ possibly being larger than $r$.\\
As detailed in the appendix, the case without sum-to-one constraints on $A$
can be handled similarly, as can be the model in \cite{Meeds2007} with binary left and right
factor and real-valued middle factor.  
\\
\textbf{Computational complexity.} The dominating cost in Algorithm
\ref{alg:1} is computation of the candidate matrix $\wh{T}$ and checking
whether its columns are vertices of $[0,1]^m$. Note that
\begin{equation}\label{eq:omitrows}
\wh{T}_{\mc{R},:} = Z_{\mc{R},:} (B^{(r-1)}  -
  p_{\mc{R}} \bm{1}_{2^{r-1}}^{\T}) + p_{\mc{R}} \bm{1}_{2^{r-1}}^{\T} =
  I_{r-1}(B^{(r-1)}  - p_{\mc{R}} \bm{1}_{2^{r-1}}^{\T})  + p_{\mc{R}} \bm{1}_{2^{r-1}}^{\T} = B^{(r-1)},
\end{equation}       
i.e.~the $r-1$ rows of $\wh{T}$ corresponding to $\mc{R}$ do not need to be
taken into account. Forming the matrix $\wh{T}$ would hence require
$O((m-r+1) (r-1) 2^{r-1})$ and the subsequent check for vertices in
the fourth step $O((m-r+1) 2^{r-1})$ operations. All other operations are of lower
order provided e.g.~$(m - r + 1) 2^{r-1} > n$. The second most expensive
operation is forming the matrix $P_{\mc{R},\mc{C}}$ in step 2 with the help of
a QR decomposition requiring $O(m n (r-1))$ operations in typical cases \cite{Gu1996}. Computing the matrix
factorization \eqref{eq:exactfactorize} after the vertices have been
identified (steps 2 to 4 in Algorithm \ref{alg:2}) has complexity $O(m n r +
r^3 + r^2 n)$. Here, the dominating part is the solution of a linear system in
$r$ variables and $n$ right hand sides. Altogether, our approach
for solving \eqref{eq:exactfactorize} has exponential complexity in $r$,
but only linear complexity in $m$ and $n$. Later on, we will argue that under
additional assumptions on $T$, the $O((m - r + 1) 2^{r-1})$ terms can be
reduced to $O((r-1) 2^{r-1})$.\\      
\\
\textbf{2.3 Uniqueness.} 
In this section, we study uniqueness of the matrix factorization
problem \eqref{eq:exactfactorize} (modulo permutation of columns/rows). First note that in
view of the affine independence of the columns of $T$, the factorization is
unique iff $T$ is, which holds iff
\begin{equation}\label{eq:uniquenessbasic}
\aff(D) \cap \{0,1\}^m = \aff(T) \cap \{0,1\}^m = \{T_{:,1},\ldots,T_{:,r} \},
\end{equation}
i.e.~if the affine subspace generated by $\{ T_{:,1},\ldots,T_{:,r} \}$
contains no other vertices of $[0,1]^m$ than the $r$ given ones (cf.~Figure 1). Uniqueness is of
great importance in applications, where one aims at an interpretation in which the columns of $T$ play the role of underlying
data-generating elements. Such an interpretation is not valid if
\eqref{eq:uniquenessbasic} fails to hold, since it is then possible to replace one
of the columns of a specific choice of $T$ by another vertex contained in the
same affine subspace.\\ 
\textbf{Solution of a non-negative variant of our factorization.} 
In the sequel, we argue that property \eqref{eq:uniquenessbasic} plays an
important role from a computational point of view when solving extensions of problem \eqref{eq:exactfactorize} in
which further constraints are imposed on $A$. One
particularly important extension is the following.
\begin{equation} \label{eq:exactfactorize_nmf}
\text{find} \; T \in \{0,1\}^{m \times r} \; \, \text{and} \; \, A \in \R_+^{r
  \times n}, \, A^{\T} \bm{1}_r = \bm{1}_n \; \; \text{such that} \; \, D = TA.
\end{equation}
Problem \eqref{eq:exactfactorize_nmf} is a special instance of non-negative
matrix factorization. Problem \eqref{eq:exactfactorize_nmf} is of particular
interest in the present paper, leading to a novel real world application of
matrix factorization techniques as presented in Section 4.2 below. It is natural to ask whether Algorithm \ref{alg:2} can be adapted to solve 
problem \eqref{eq:exactfactorize_nmf}. A change is obviously required for the second
step when selecting $r$ vertices from $\mc{T}$, since in
\eqref{eq:exactfactorize_nmf} the columns $D$ now have to be expressed as convex instead of only
affine combinations of columns of $T$: picking an affinely
independent collection from $\mc{T}$ does not take into account the
non-negativity constraint imposed on $A$. If, however,
\eqref{eq:uniquenessbasic} holds, we have $|\mc{T}| = r$ and Algorithm
\ref{alg:2} must return a solution of \eqref{eq:exactfactorize_nmf} provided
that there exists one.
\begin{corro}\label{prop:nmf} If problem \eqref{eq:exactfactorize} has a unique solution,
i.e.~if condition \eqref{eq:uniquenessbasic} holds and if there exists a solution of \eqref{eq:exactfactorize_nmf}, then it is returned
by Algorithm \ref{alg:2}.   
\end{corro}
To appreciate that result, consider the converse case $|\mc{T}| > r$. 
Since the aim is a minimal factorization, one has to find a subset of $\mc{T}$ of cardinality
$r$ such that \eqref{eq:exactfactorize_nmf} can be solved. In principle, this
can be achieved by solving a linear program for
$\binom{|\mc{T}|}{r}$ subsets of $\mc{T}$, but this is in general not computationally feasible: 
the upper bound of Proposition \ref{prop:basic} indicates that
$|\mc{T}| = 2^{r-1}$ in the worst case. For the example below, $\mc{T}$ consists of all $2^{r-1}$ vertices contained in an
$r-1$-dimensional face of $[0,1]^m$:
\begin{equation}\label{eq:badT}
T = {\scriptsize \begin{pmatrix}
     0_{m-r \times r} \\
 \\
    I_{r-1} \; 0_{r-1} \\
\\
     0_{r}^{\T} 
    \end{pmatrix}} \quad \text{with} \; \mc{T} = \left \{T \lambda:\; \lambda_1 \in
  \{0,1\},\ldots,\lambda_{r-1} \in \{0,1\}, \, \lambda_r = 1 - \sum_{k =
    1}^{r-1} \lambda_k  \right \}. 
\end{equation}

\textbf{Uniqueness under separability.} 
In view of the negative example \eqref{eq:badT}, one might ask whether uniqueness
according to \eqref{eq:uniquenessbasic} can be achieved under
additional conditions on $T$. We prove uniqueness under \emph{separability}, a condition introduced in \cite{DonohoStodden03}
and imposed recently in \cite{Arora2012} to show solvability of the NMF problem by linear
programming. We say that $T$ is separable if there exists a permutation
$\Pi$ such that $\Pi T = [M; I_r]$, where $M \in \{0,1 \}^{m-r \times r}$.
\begin{prop}\label{prop:separability} If $T$ is separable, condition \eqref{eq:uniquenessbasic} holds
and thus problem \eqref{eq:exactfactorize} has a
unique solution.  
\end{prop}

\textbf{Uniqueness under generic random sampling.} Both the negative example \eqref{eq:badT} as well as the positive result of
Proposition \ref{prop:separability} are associated with special
matrices $T$. This raises the question whether uniqueness holds respectively
fails for broader classes of binary matrices. In order to gain insight
into this question, we consider random $T$ with i.i.d.~entries from a
Bernoulli distribution with parameter $\frac{1}{2}$
and study the probability of the event $\{ \aff(T) \cap \{0,1\}^m =
\{T_{:,1},\ldots,T_{:,r} \} \}$. This question has essentially been studied in
combinatorics \cite{Odlyzko1988}, with further improvements in
\cite{Kahn1995}. The results therein rely crucially on Littlewood-Offord theory (see Section
2.4 below). 
\begin{theo}\label{corro:kahn} Let $T$ be a random $m \times r$-matrix 
whose entries are drawn i.i.d.~from $\{0,1\}$ with probability
$\frac{1}{2}$. Then, there is a constant
  $C$ so that if $r \leq m-C$,
\begin{equation*} 
\p \Big( \aff(T) \cap \{0,1\}^m =
\{T_{:,1},\ldots,T_{:,r} \Big) \geq 1 - (1
+ o(1)) \, 4 \binom{r}{3} \left( \frac{3}{4} \right)^m - \left( \frac{3}{4} +
  o(1) \right)^m\; \, \text{as} \; m
\rightarrow \infty. 
\end{equation*}
\end{theo}
Theorem \ref{corro:kahn} suggests a positive answer to the question of
uniqueness posed above. For $m$ large enough and $r$ small
compared to $m$ (in fact, following \cite{Kahn1995} one may conjecture that Theorem
\ref{corro:kahn} holds with $C = 1$), the probability that the affine hull of $r$ vertices of $[0,1]^m$ selected uniformly at random contains
some other vertex is exponentially small in the dimension $m$. We have
empirical evidence that the result of Theorem \ref{corro:kahn} continues to
hold if the entries of $T$ are drawn from a Bernoulli distribution with
parameter in $(0,1)$ sufficiently far away from the boundary points (cf.~appendix). As a byproduct, these results imply that also the NMF variant of our matrix factorization
problem \eqref{eq:exactfactorize_nmf} can in most cases be reduced to identifying a set of
$r$ vertices of $[0,1]^m$ (cf.~Corollary \ref{prop:nmf}).   

\begin{figure}
\begin{center}
\begin{tabular}{lll}
\hspace{-2cm} \includegraphics[height = 0.16\textheight]{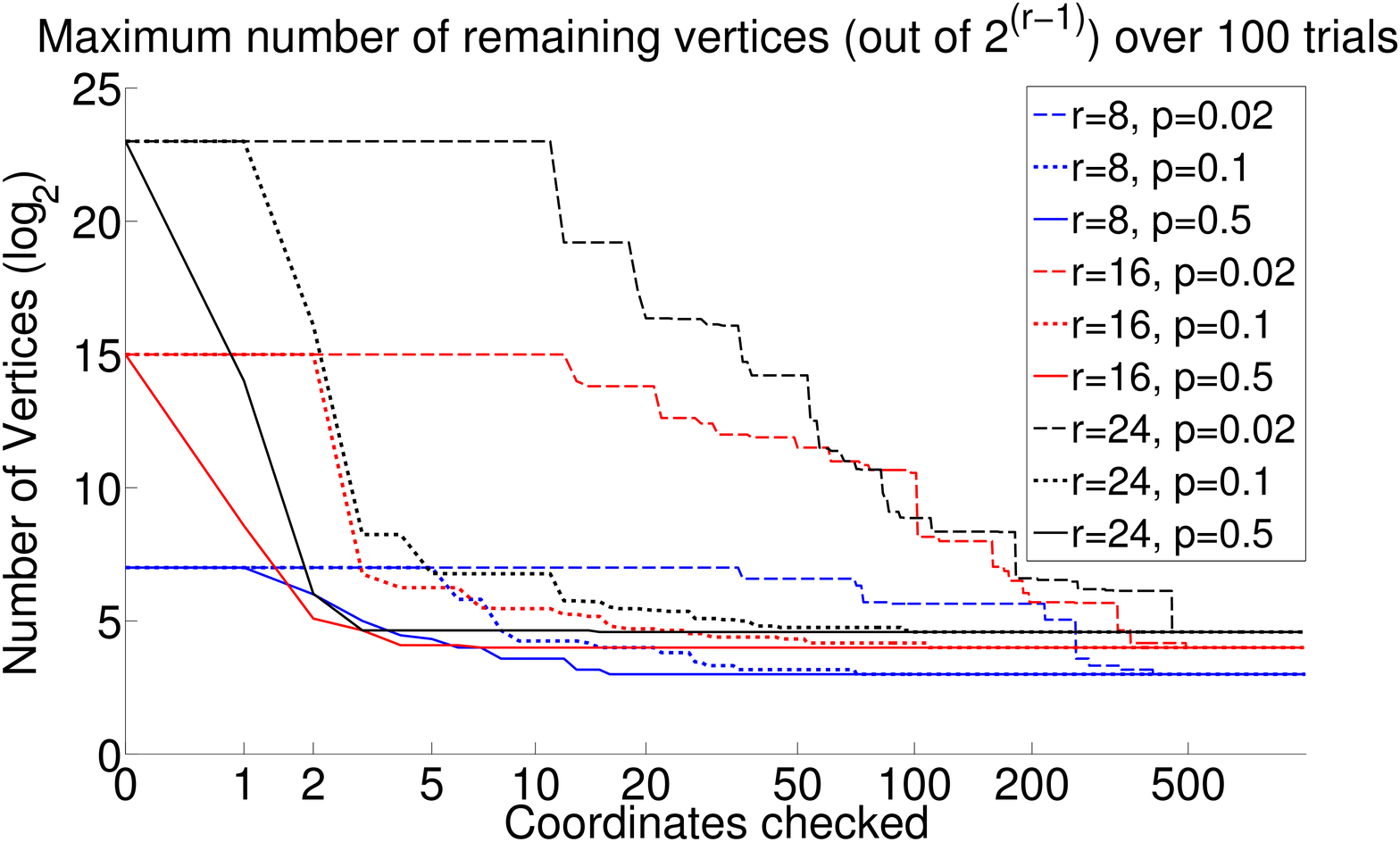}
& \hspace{0.5cm}\includegraphics[height =0.16\textheight]{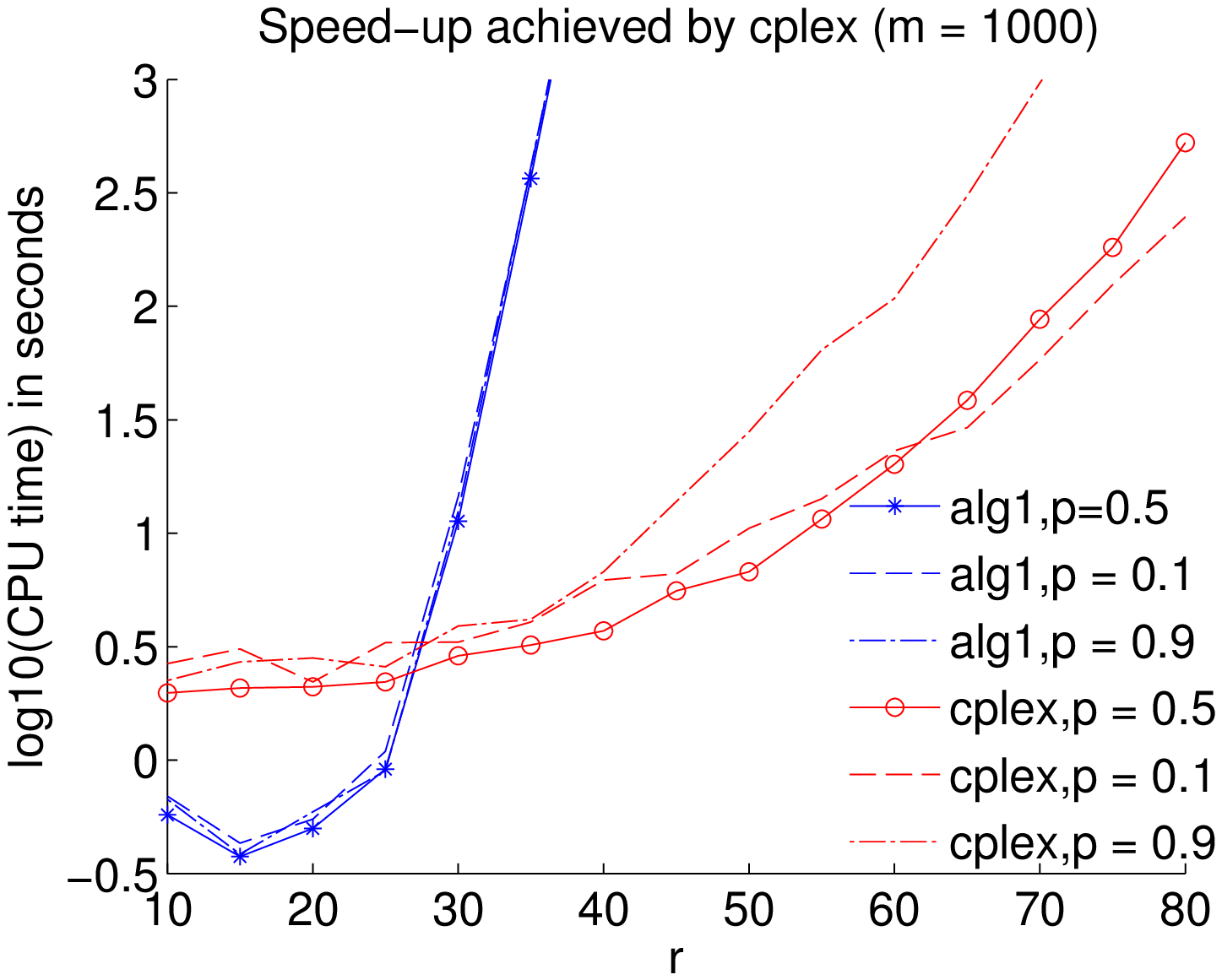}
\end{tabular}
\caption{Left: Speeding up the algorithm by checking single coordinates,
remaining number of coordinates vs.$\#$ coordinates checked ($m =
1000$). Right: Speed up by \textsf{CPLEX} compared to Algorithm
\ref{alg:1}. For both plots, $T$ is drawn entry-wise from a Bernoulli
distribution with parameter $p$.}\label{fig:speedup}
\end{center}
\end{figure}

\textbf{2.4 Speeding up Algorithm \ref{alg:1}.} 
In Algorithm \ref{alg:1}, an $m \times 2^{r-1}$ matrix $\wh{T}$ of
potential vertices is formed (Step 3). We have discussed the case
\eqref{eq:badT} where all candidates must indeed be vertices, in which case it seems to be impossible
to reduce the computational cost of $O((m-r) r 2^{r-1})$, which becomes significant 
once $m$ is in the thousands and $r \geq 25$. On the positive side, Theorem
\ref{corro:kahn} indicates that for many instances of $T$, only $r$ out of
$2^{r-1}$ candidates are in fact vertices. In that case, noting that columns
of $\wh{T}$ cannot be vertices if a single coordinate is not in $\{0,1
\}$ (and that the vast majority of columns of $\wh{T}$ must have one
such coordinate), it is computationally more favourable to
incrementally compute subsets of rows of $\wh{T}$ and then to discard already
those columns with coordinates not in $\{0,1 \}$. We have observed
empirically that this scheme rapidly reduces the candidate set $-$ already
checking a single row of $\wh{T}$ eliminates a substantial portion (see Figure
\ref{fig:speedup}).\\
\textbf{Littlewood-Offord theory.} 
Theoretical underpinning for the last observation can be obtained from a
result in combinatorics, the \emph{Littlewood-Offord}
(L-O)-lemma. Various extensions of that result have been developed until
recently, see the survey \cite{Nguyen2012}. 
We here cite the L-O-lemma in its basic form. 
\begin{theo}\label{theo:lo}\cite{Erdos1951} Let $a_1,\ldots,a_{\ell} \in \R \setminus \{0 \}$
  and $y \in \R$.
\begin{itemize}  
\item[(i)] $\big| \{b \in \{0,1\}^{\ell}:\, \sum_{i = 1}^{\ell} a_i b_i = y\}  \big|
\leq \binom{\ell}{\lfloor \ell/2 \rfloor}$.
\item[(ii)] If $|a_i| \geq 1, \, i=1,\ldots,\ell$, $\, \big| \{b \in \{0,1\}^{\ell}:\, \sum_{i = 1}^{\ell} a_i b_i \in
(y, y+1) \}  \big| \leq \binom{\ell}{\lfloor \ell/2 \rfloor}$.  
\end{itemize}
\end{theo}
The two parts of Theorem \ref{theo:lo} are referred to as discrete
respectively continuous L-O lemma. The discrete L-O lemma provides an upper
bound on the number of $\{0,1\}$-vectors whose weighted sum with given weights
$\{a_i\}_{i=1}^{\ell}$ is equal to some given number $y$, whereas the
stronger continuous version, under a more stringent condition on the weights, upper bounds the number of $\{0,1\}$-vectors whose
weighted sum is contained in some interval $(y, y+1)$. In order to see the
relation of Theorem \ref{theo:lo} to Algorithm \ref{alg:1}, let us re-inspect
the third step of that algorithm. To obtain a reduction of candidates by
checking a single row of $\wh{T} = Z (B^{(r-1)}  -  p_{\mc{R}}
\bm{1}_{2^{r-1}}^{\T}) + p \bm{1}_{2^{r-1}}^{\T}$, pick $i \notin \mc{R}$
(recall that coordinates in $\mc{R}$ do not need to be checked, cf.~\eqref{eq:omitrows})
and $u \in \{1,\ldots,2^{r-1} \}$ arbitrary. The $u$-th candidate can be a vertex only if $\wh{T}_{i,u} \in
\{0,1\}$. The condition $\wh{T}_{i,u} = 0$ can be written as 
\begin{equation}\label{eq:lolike}
\underbrace{Z_{i,:}}_{\{ a_k \}_{k=1}^r} \underbrace{B_{:,u}^{(r-1)}}_{=b} = \underbrace{Z_{i,:} p_{\mc{R}} - p_i}_{=y}. 
\end{equation}
A similar reasoning applies when setting $\wh{T}_{i,u} = 1$. Provided none of
the entries of $Z_{i,:} = 0$, the discrete L-O lemma implies that there are at
most $2 \binom{r-1}{\lfloor (r-1)/2 \rfloor}$ out of $2^{r-1}$ candidates for
which the $i$-th coordinate is in $\{0,1\}$. This yields a
reduction of the candidate set by $2 \binom{r-1}{\lfloor (r-1)/2 \rfloor}/2^{r-1} =
O \left(\frac{1}{\sqrt{r-1}} \right)$. Admittedly, this reduction may appear
insignificant given the total number of candidates to be checked. The
reduction achieved empirically (cf.~Figure \ref{fig:speedup}) is typically
larger. 
Stronger reductions have been proven under 
additional assumptions on the weights $\{a_i \}_{i=1}^{\ell}$: e.g.~for distinct
weights, one obtains a reduction of $O((r-1)^{-3/2})$
\cite{Nguyen2012}. Furthermore, when picking successively $d$ rows of
$\wh{T}$ and if one assumes that each row yields a reduction according to the
discrete L-O lemma, one would obtain the reduction $(r-1)^{-d/2}$ so that $d =
r-1$ would suffice to identify all vertices provided $r \geq 4$. Evidence for
the rate $(r-1)^{-d/2}$ can be found in \cite{Tao2012}. This indicates a reduction in complexity of Algorithm
\ref{alg:1} from $O((m-r) r 2^{r-1})$ to $O(r^2 2^{r-1})$.
 \hfill \\ \textbf{Achieving further speed-up with integer linear
   programming.}$\,$
 The continuous L-O lemma (part (ii) of Theorem \ref{theo:lo}) combined with the
derivation leading to \eqref{eq:lolike} 
allows us to tackle even the case $r = 80$ ($2^{80} \approx 10^{24}$). In
view of the continuous L-O lemma, a reduction in the number of candidates can still
be achieved if the requirement is weakened to $\wh{T}_{i,u} \in
[0,1]$. According to \eqref{eq:lolike} the candidates satisfying the relaxed constraint for the $i$-th
coordinate can be obtained from the feasibility problem  
\begin{equation}\label{eq:ilp}
\text{find} \; b \in \{0,1\}^{r-1} \; \, \text{subject to} \; 0 \leq  Z_{i,:} (b - p_{\mc{R}}) + p_i\leq 1,
\end{equation}
which is an integer linear program that can be solved e.g.~by \textsf{CPLEX}. The L-O- theory suggests that the branch-bound
strategy employed therein is likely to be successful. With the help of
\textsf{CPLEX}, it is affordable to solve problem \eqref{eq:ilp} with all $m-r+1$ constraints (one for each of the rows of
$\wh{T}$ to be checked) imposed simultaneously. 
We always recovered directly the underlying vertices in our
experiments and only these, without the need to prune the solution
pool (which could be achieved by Algorithm 1, replacing the 
$2^{r-1}$ candidates by a potentially much smaller solution pool).
\section{Approximate case}\label{sec:approximate}
\vspace{-0.15cm}
In the sequel, we discuss an extension of our approach to handle the approximate case  
$D \approx T A$ with $T$ and $A$ as in
\eqref{eq:exactfactorize}. In particular, we have in mind the case of additive
noise i.e.~$D = T A + E$ with $\nnorm{E}_F$ small. While the basic concept of Algorithm \ref{alg:1}
can be adopted, changes are necessary because $D$ may have full rank
$\min\{m,n\}$ and second $\aff(D) \cap \{0,1 \}^m = \emptyset$, i.e.~the
distances of $\aff(D)$ and the $\{ T_{:,k} \}_{k = 1}^r$ may be strictly
positive (but are at least assumed to be small).
\vspace{-0.25cm}    
\begin{algorithm}
\caption{\textsc{FindVertices} \textsc{approximate}}\label{alg:3}
\begin{algorithmic}
\STATE \begin{enumerate}\item[1.] Let $p = D \bm{1}_n/n$ and compute $P = [D\cls{1} -p,\ldots,D\cls{n} - p]$.
     \end{enumerate}   
\STATE \begin{enumerate}\item[2.] Compute $U^{(r - 1)}  \in \R^{m
    \times r-1}$, the left singular vectors corresponding to the $r-1$ largest singular values of
  $P$. Select $r-1$ linearly independent rows $\mc{R}$ of $U^{(r-1)}$,
  obtaining $U_{\mc{R},:}^{(r-1)} \in \R^{r-1 \times r-1}$. 
\end{enumerate}
\STATE \begin{enumerate}\item[3.] Form $Z =  U^{(r-1)} (U_{\mc{R},:}^{(r-1)})^{-1}$ and $\wh{T} = Z (B^{(r-1)}  -
  p_{\mc{R}} \bm{1}_{2^{r-1}}^{\T}) + p \bm{1}_{2^{r-1}}^{\T}$. \end{enumerate}
\STATE \begin{enumerate}\item[4.] Compute $\wh{T}^{01} \in \R^{m \times
    2^{r-1}}$: for $u =  1,\ldots,2^{r-1}$, $i =
  1,\ldots,m$, set $\wh{T}_{i,u}^{01} = I(\wh{T}_{i,u} > \frac{1}{2})$.
\end{enumerate}
\STATE  \begin{enumerate}\item[5.] For $u=1,\ldots,2^{r-1}$, set $\delta_u =
  \nnorm{\wh{T}\cls{u} - \wh{T}\cls{u}^{01}}_2$. Order increasingly s.t. $\delta_{u_1}
  \leq \ldots \leq \delta_{2^{r-1}}$.\end{enumerate}
\STATE \begin{enumerate}\item[6.] Return $T = [\wh{T}\cls{u_1}^{01} \ldots \wh{T}\cls{u_r}^{01}]$ 
\end{enumerate}
\end{algorithmic}
\end{algorithm}
As distinguished from the exact case, Algorithm \ref{alg:3} requires
the number of components $r$ to be specified in advance as it is typically the
case in noisy matrix factorization problems. Moreover, the vector $p$
subtracted from all columns of $D$ in step 1 is chosen as the mean of the data
points, which is in particular a reasonable choice if $D$ is
contaminated with additive noise distributed symmetrically around zero. The
truncated SVD of step 2 achieves the desired dimension reduction and potentially reduces noise corresponding to small singular values that are
discarded. The last change arises in step 5. While in the exact case, one identifies all columns of $\wh{T}$ that are in
$\{ 0,1 \}^m$, one instead only identifies columns close to $\{0,1\}^m$. Given
the output of Algorithm \ref{alg:3}, we solve the approximate matrix
factorization problem via least squares, obtaining the right factor from $\min_{A} \nnorm{D - TA}_F^2$.\\          
\textbf{Refinements.} Improved performance for higher noise levels can be achieved by running
Algorithm \ref{alg:3} multiple times with different sets of rows selected in
step 2, which yields candidate matrices $\{ T^{(l)} \}_{l=1}^s$, and
subsequently using $T = \argmin_{\{ T^{(l)} \} }\min_{A} \nnorm{D - T^{(l)} A}_F^2$, i.e.~one picks the candidate yielding the best fit.  
Alternatively, we may form a candidate pool by merging the $\{ T^{(l)}
\}_{l=1}^s$  and then use a backward elimination scheme, in which successively
candidates are dropped that yield the smallest improvement in fitting $D$
until $r$ candidates are left. Apart from that, $T$ returned
by Algorithm \ref{alg:3} can be used for initializing the block
optimization scheme of Algorithm \ref{alg:4} below.
\begin{algorithm}
\caption{Block optimization scheme for solving $\min_{T \in \{0,1\}^{m \times r}, \, A} \,\nnorm{D - TA}_F^2$}\label{alg:4}
\begin{algorithmic}
\STATE
\hspace{-0.3cm}1. Set $k = 0$ and set $T^{(k)}$ equal to a starting value.\\
\hspace{-0.3cm}2. $A^{(k)} \leftarrow \argmin_{A} \nnorm{D - T^{(k)} A}_F^2$ and set $k = k+1$.\\
\hspace{-0.3cm}3. $T^{(k)} \leftarrow \argmin_{T \in
    \{0,1\}^{m \times r}} \nnorm{D - T A^{(k)}}_F^2 =  \argmin_{\{T_{i,:} \in
    \{0,1\}^r\}_{i = 1}^m} \sum_{i=1}^m  \nnorm{D_{i,:}-T_{i,:}
    A^{(k)}}_2^2 \,\text{(9)}$\\
\hspace{-0.3cm}4. Alternate between steps 2 and 3.
\end{algorithmic}
\end{algorithm} 
Algorithm \ref{alg:4} is akin to standard block coordinate descent schemes
proposed in the matrix factorization literature, e.g.~\cite{Lin2007}. An
important observation (step 3) is that optimization of $T$ is separable along the rows
of $T$, so that for small $r$, it is feasible to perform exhaustive search over
all $2^r$ possibilities (or to use \textsf{CPLEX}). However,
Algorithm \ref{alg:4} is impractical as a stand-alone scheme, because without
proper initialization, it may take many iterations to converge,
with each single iteration being more expensive than Algorithm
\ref{alg:3}. When initialized with the output of the latter, however, we have observed
convergence of the block scheme only after few steps.

\section{Experiments}\label{sec:experiments}
\vspace{-0.15cm}
In Section 4.1 we demonstrate with the help of synthetic data that the
approach of Section \ref{sec:approximate} performs well on noisy datasets. In the second
part, we present an application to a real dataset.\\
\\  
\textbf{4.1 Synthetic data.}\\
\textbf{Setup.} We generate $D = T^* A^* + \alpha E$, where the entries of
$T^*$ are drawn i.i.d.~from $\{0,1\}$ with probability 0.5, the columns of $A$
are drawn i.i.d.~uniformly from the probability simplex and the entries of $E$
are i.i.d.~standard Gaussian. We let $m = 1000$, $r=10$ and $n =
2r$ and let the noise level $\alpha$ vary along a grid starting from 0. Small sample sizes $n$   
as considered here yield more challenging problems and are motivated by the
real world application of the next subsection.\\
\textbf{Evaluation.} Each setup is run 20 times and we report averages
over the following performance measures: the normalized Hamming distance
$\nnorm{T^* - T}_F^2 / (m\,r)$ and the two RMSEs $\nnorm{T^*A^* - TA}_F/(m \,
n)^{1/2}$ and $\nnorm{T A - D}_F/(m \, n)^{1/2}$, where $(T,A)$ denotes the
output of one of the following approaches that are
compared. \textsf{FindVertices}: our approach in Section
\ref{sec:approximate}. \textsf{oracle}: we solve problem (9) with
$A^{(k)} = A^*$. \textsf{box}: we
run the block scheme of Algorithm \ref{alg:4}, relaxing the integer constraint
into a box constraint. Five random initializations are used and we take the
result yielding the best fit, subsequently rounding the entries of $T$ to
fulfill the $\{0,1\}$-constraints and refitting $A$. \textsf{quad pen}: as
\textsf{box}, but a (concave) quadratic penalty $\lambda \sum_{i,k} T_{i,k} (1 -
T_{i,k})$ is added to push the entries of $T$ towards
$\{0,1\}$. D.C.~programming \cite{DCprogramming} is used for the block updates
of $T$.       
\begin{figure}[h!]
\begin{flushleft}
\begin{tabular}{lll}
\hspace{-0.28cm}\includegraphics[height = 0.15\textheight]{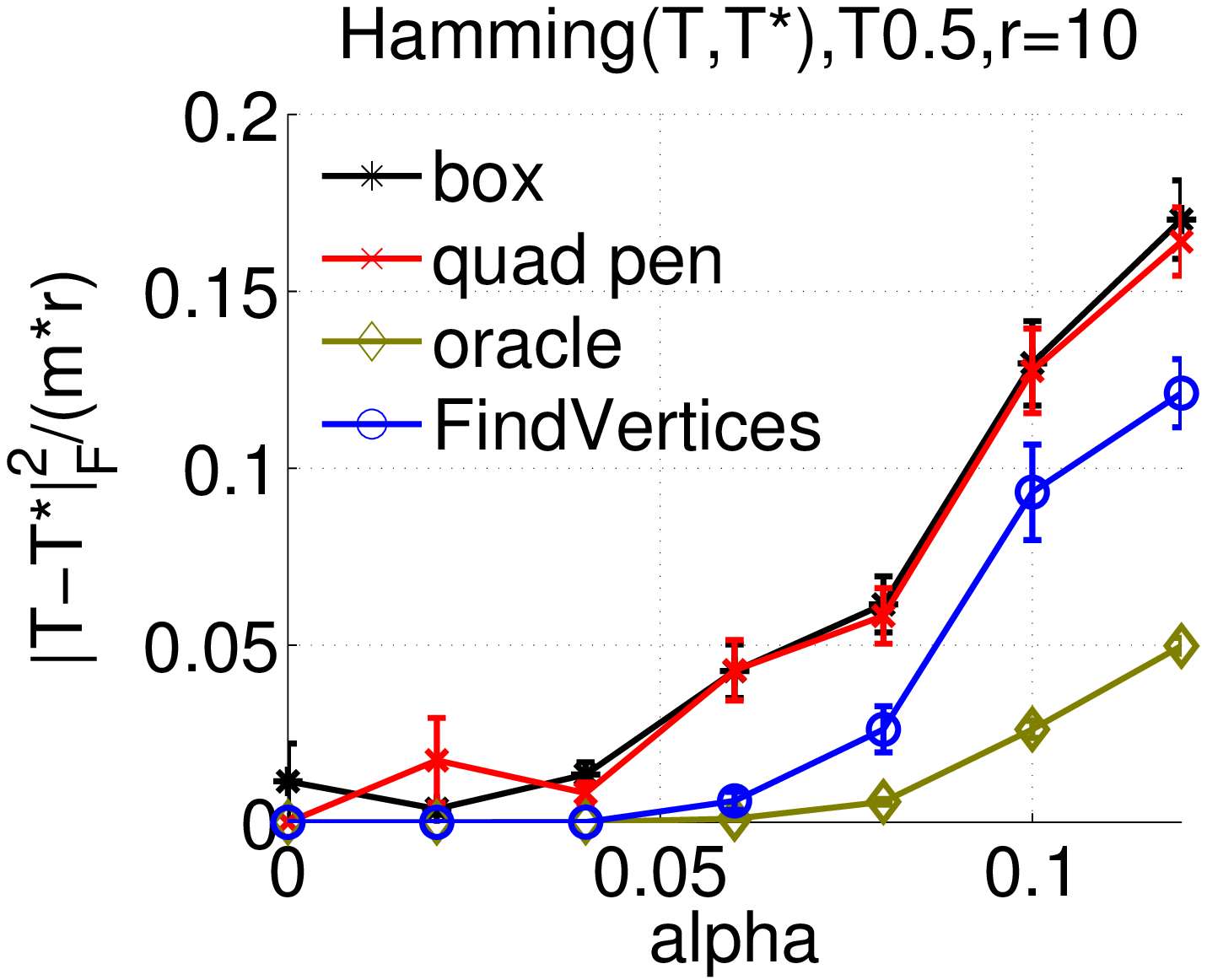}
&\hspace{-0.28cm} \includegraphics[height = 0.15\textheight]{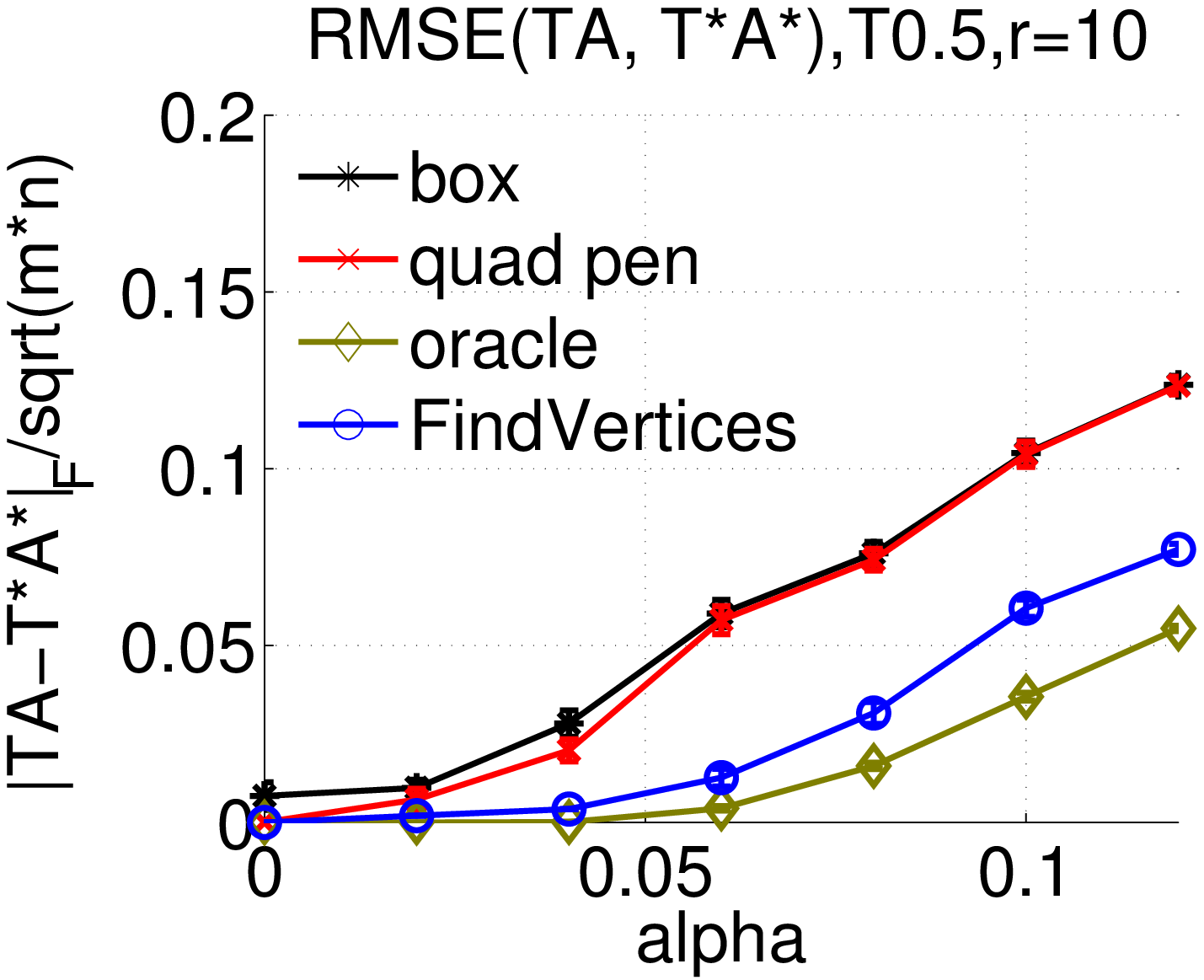}
&\hspace{-0.29cm} \includegraphics[height =
0.15\textheight]{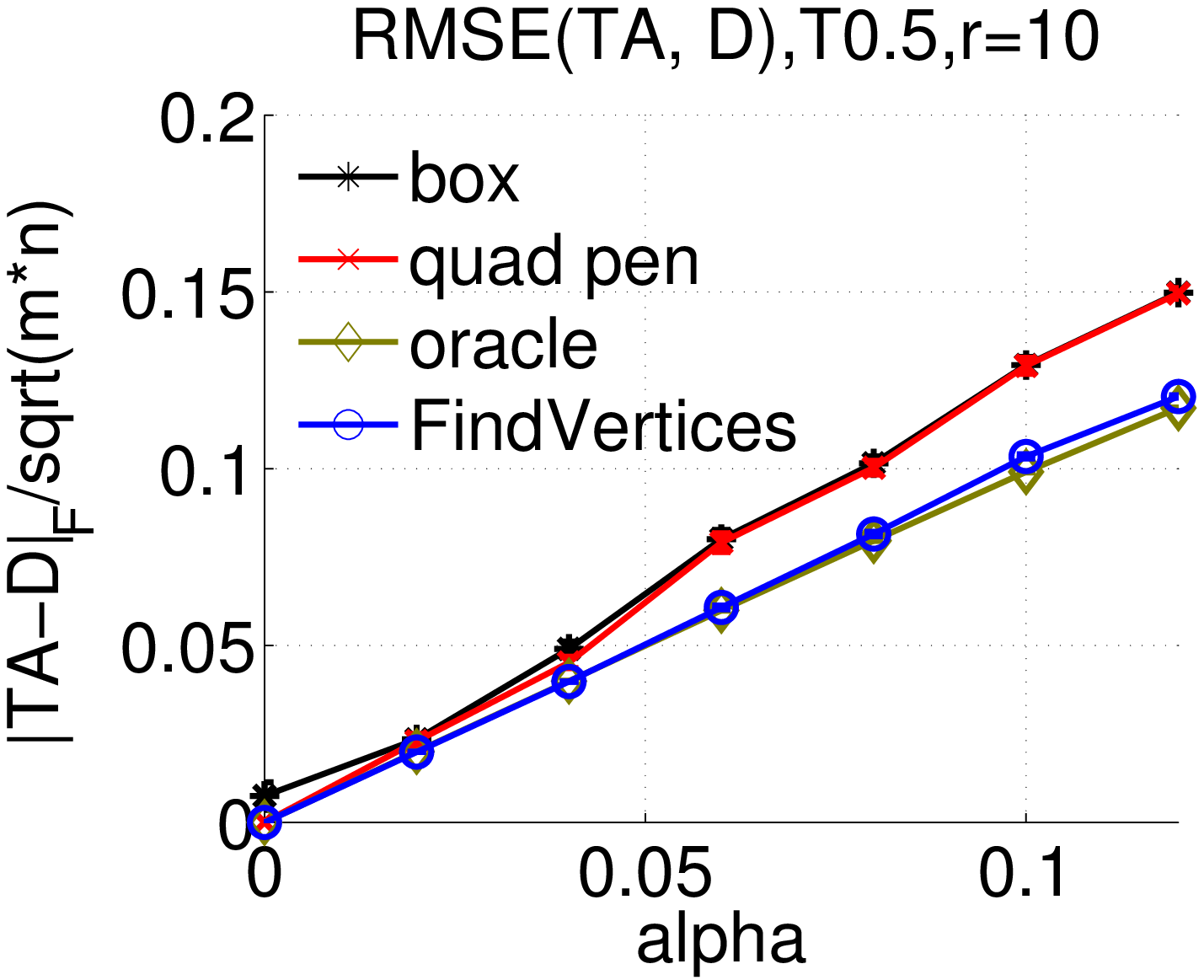}\\
\hspace{-0.28cm}\includegraphics[height = 0.15\textheight]{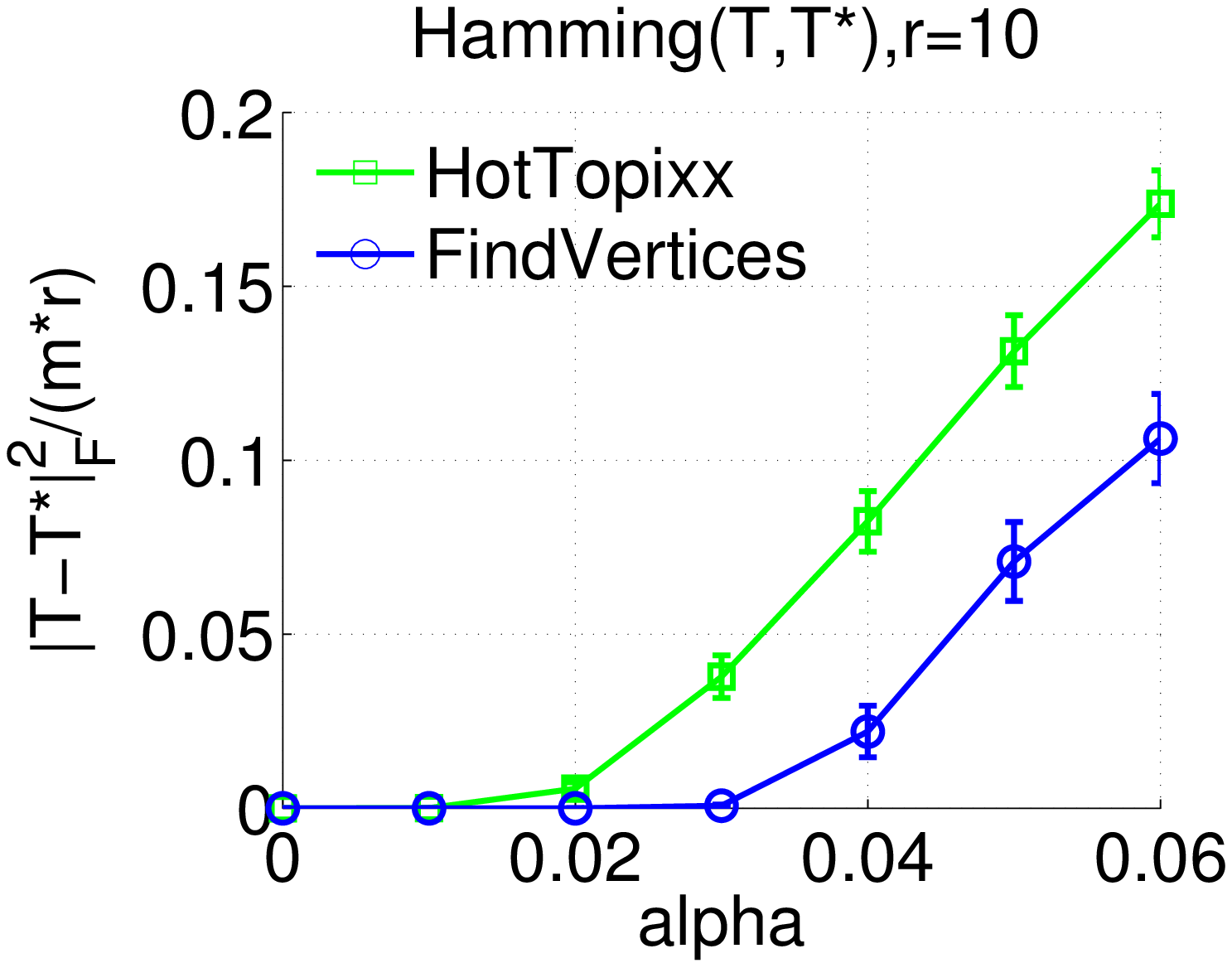}
&\hspace{-0.28cm} \includegraphics[height = 0.15\textheight]{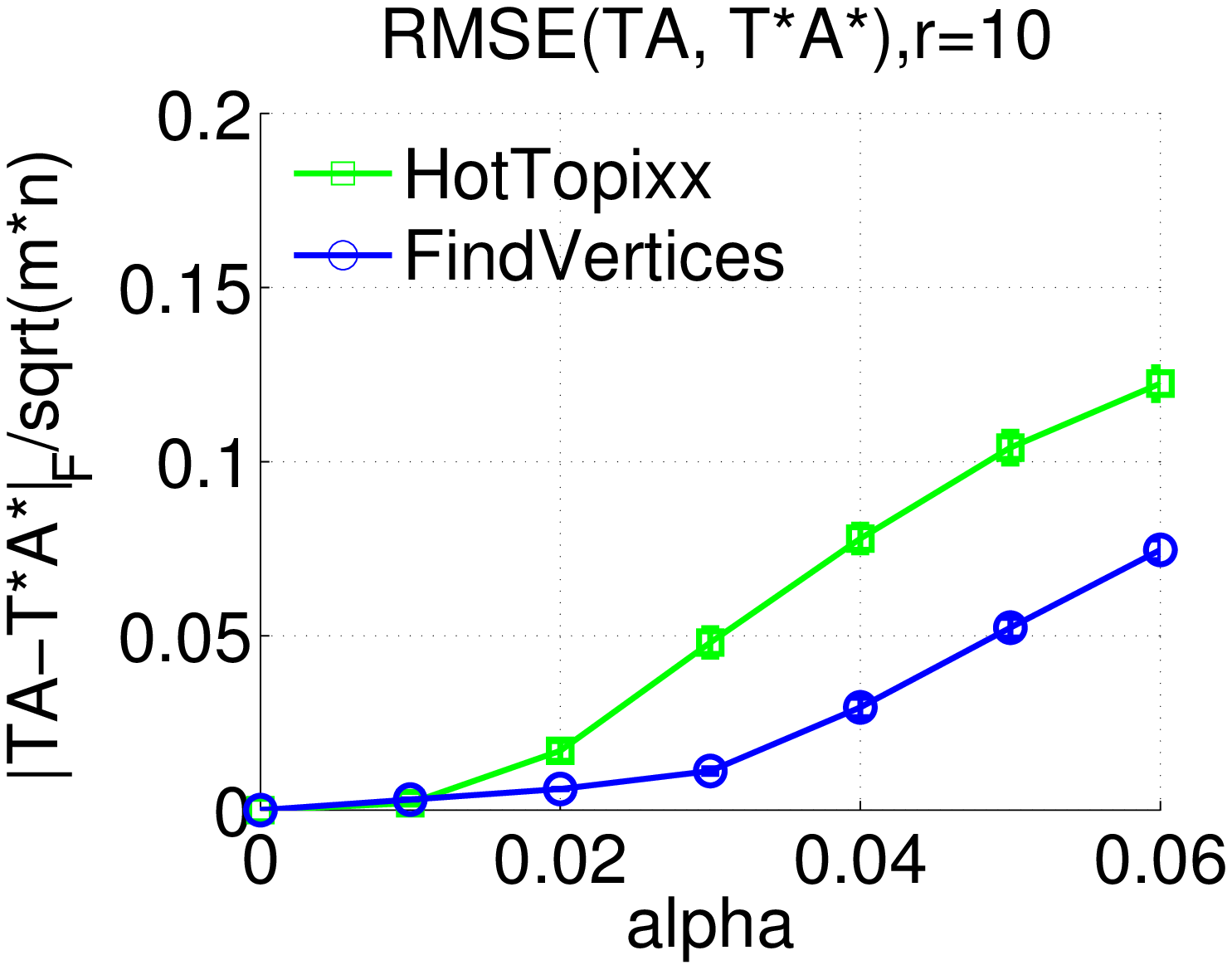}
&\hspace{-0.29cm} \includegraphics[height =
0.15\textheight]{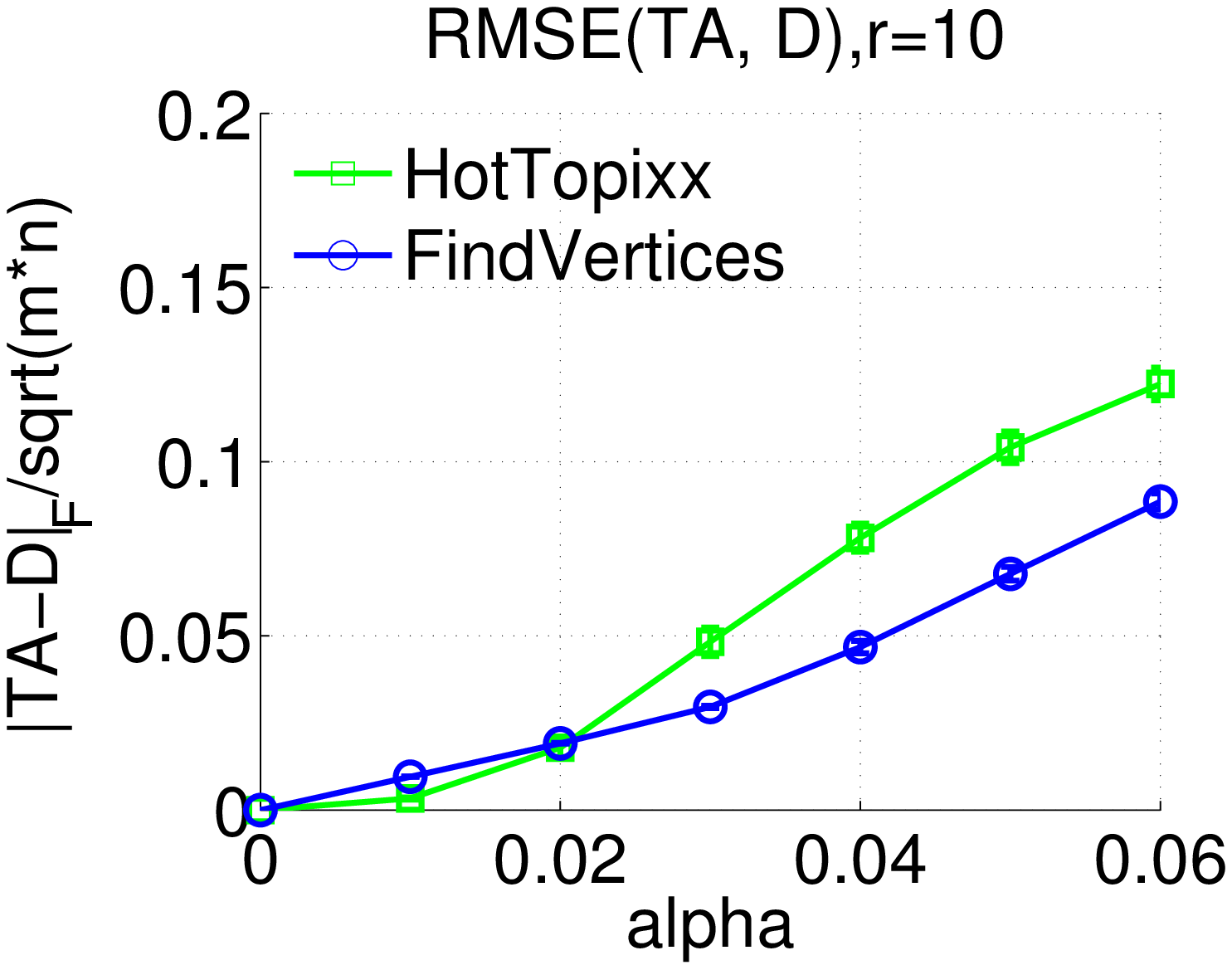}

\end{tabular}
\caption{Top: comparison against
  block schemes. Bottom: comparison against \textsf{HOTTOPIXX}. Left/Middle/Right: $\nnorm{T^* - T}_F^2 / (m\,r)$, $\nnorm{T^*A^* - TA}_F/(m \,
n)^{1/2}$ and $\nnorm{T A - D}_F/(m \, n)^{1/2}$.}\label{fig:toy1}
\end{flushleft}
\end{figure} \hfill \\
\textbf{Comparison to \textsf{HOTTOPIXX} \cite{Bittdorf2012}.} \textsf{HOTTOPIXX} (\textsf{HT}) is a linear
programming approach to NMF equipped with guarantees such as correctness in the exact and robustness in the non-exact case as long as $T$
is (nearly) separable (cf.~Section 2.3). \textsf{HT}
does not require $T$ to be binary, but applies to the generic NMF problem $D
\approx T A$, $T \in \R_{+}^{m \times r}$ and $A \in \R_{+}^{r \times
  n}$. Since separability is crucial to the performance of \textsf{HT},
we restrict our comparison to separable $T = [M;I_r]$, generating the entries
of $M$ i.i.d.~from a Bernoulli distribution with parameter $0.5$. For runtime reasons, we lower the dimension to $m=100$. Apart from
that, the experimental setup is as above. We use an implementation of
\textsf{HT} from \cite{gilliscode}. We first pre-normalize $D$ to have unit row sums as
required by \textsf{HT}, and obtain $A$ as first output. Given $A$, the non-negative least
squares problem $\min_{T \in \R_+^{m \times r}} \nnorm{D - T A}_F^2$ is
solved. The entries of $T$ are then re-scaled to match the original
scale of $D$, and thresholding at 0.5 is applied to obtain a binary
matrix. Finally, $A$ is re-optimized by solving the above fitting problem with
respect to $A$ in place of $T$. In the noisy case, \textsf{HT} needs a
tuning parameter to be specified that depends on the noise level, and we
consider a grid of $12$ values for that parameter. The range of the grid is
chosen based on knowledge of the noise matrix $E$. For
each run, we pick the parameter that yields best performance in favour of
\textsf{HT}.\\
\textbf{Results.} From Figure \ref{fig:toy1}, we find that unlike the other
approaches, \textsf{box} does not always recover $T^*$ even if
the noise level $\alpha = 0$. \textsf{FindVertices} outperforms \textsf{box}
and \textsf{quad pen} throughout. For $\alpha \leq 0.06$, its performance 
closely matches that of the oracle. In the separable
case, our approach performs favourably as compared to \textsf{HT}, a
natural benchmark in this setting.\\ 
\\
\textbf{4.2 Analysis of DNA methylation data.}\\
\textbf{Background.} Unmixing of DNA methylation profiles is a problem of high
interest in cancer research. DNA methylation is a chemical modification of
the DNA occurring at specific sites, so-called CpGs. DNA methylation affects
gene expression and in turn various processes such as cellular
differentiation. A site is either unmethylated ('0') or methylated ('1'). DNA
methylation microarrays allow one to measure the methylation level for
thousands of sites. In the
dataset considered here, the measurements $D$ (the rows corresponding to
sites, the columns to samples) result from a mixture of
cell types. The methylation profiles of the latter are in $\{0,1\}^m$,
whereas, depending on the mixture proportions associated with each sample, the
entries of $D$ take values in $[0,1]^m$. In other words, we have the model $D
\approx T A$, with $T$ representing the methylation of the cell types
and the columns of $A$ being elements of the probability simplex. It is often of interest to recover the mixture
proportions of the samples, because e.g.~specific diseases, in particular
cancer, can be associated with shifts in these proportions. The matrix $T$ is
frequently unknown, and determining it experimentally is costly. Without $T$,
however, recovering the mixing matrix $A$ is challenging, in particular
since the number of samples in typical studies is small.\\         
\textbf{Dataset.} We consider the dataset studied in \cite{Houseman2012short},
with $m = 500$ CpG sites and $n = 12$ samples of blood cells composed of four
major types (B-/T-cells, granulocytes, monocytes), i.e.~$r=4$. Ground
truth is partially available: the proportions of the samples, denoted by
$A^*$, are known. \vspace{-0.35cm}
\begin{figure}[h!] 
\begin{flushleft}
\begin{tabular}{lll}
\hspace{-0.28cm}\includegraphics[height = 0.16\textheight]{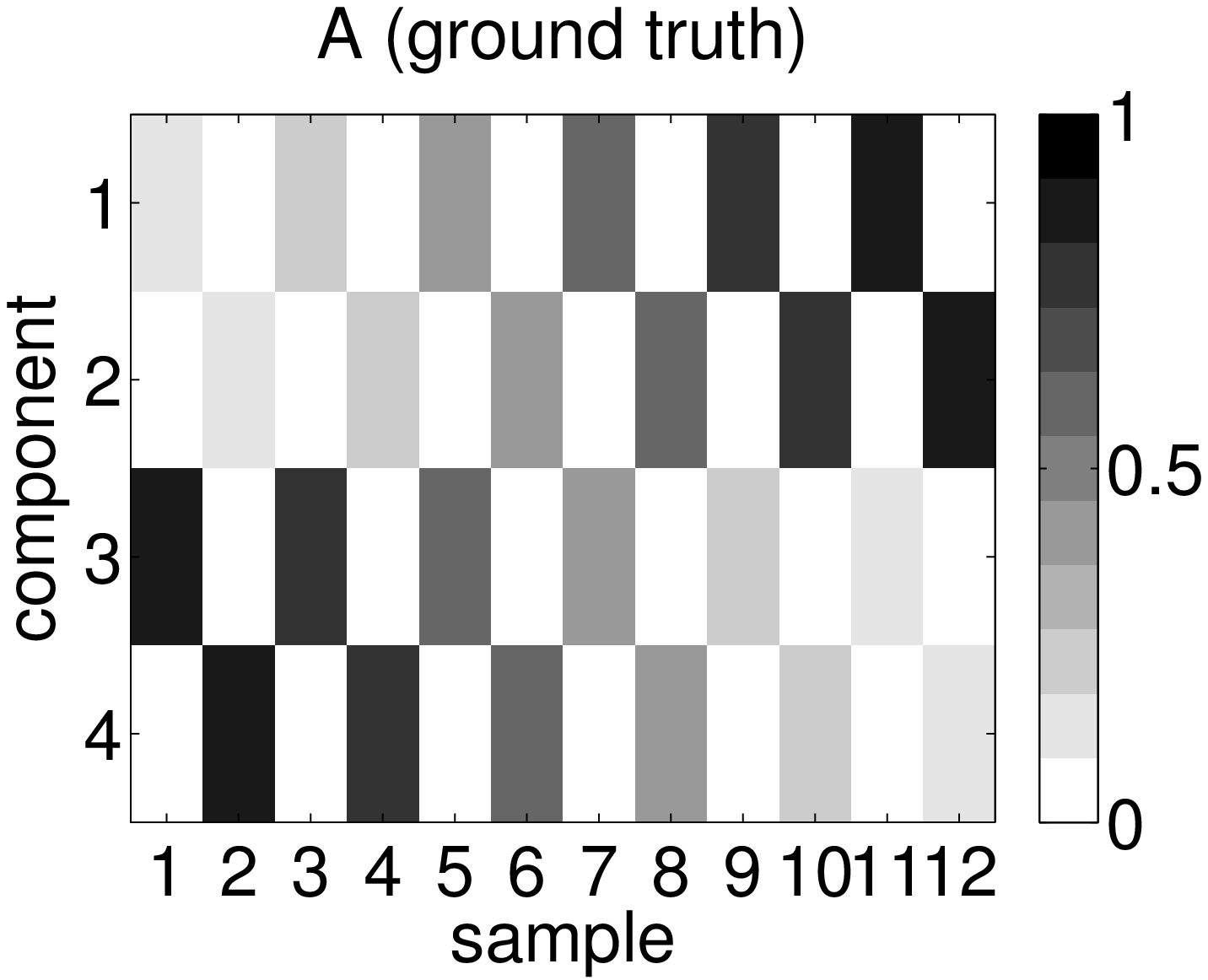}
&\hspace{-0.40cm} \includegraphics[height = 0.16\textheight]{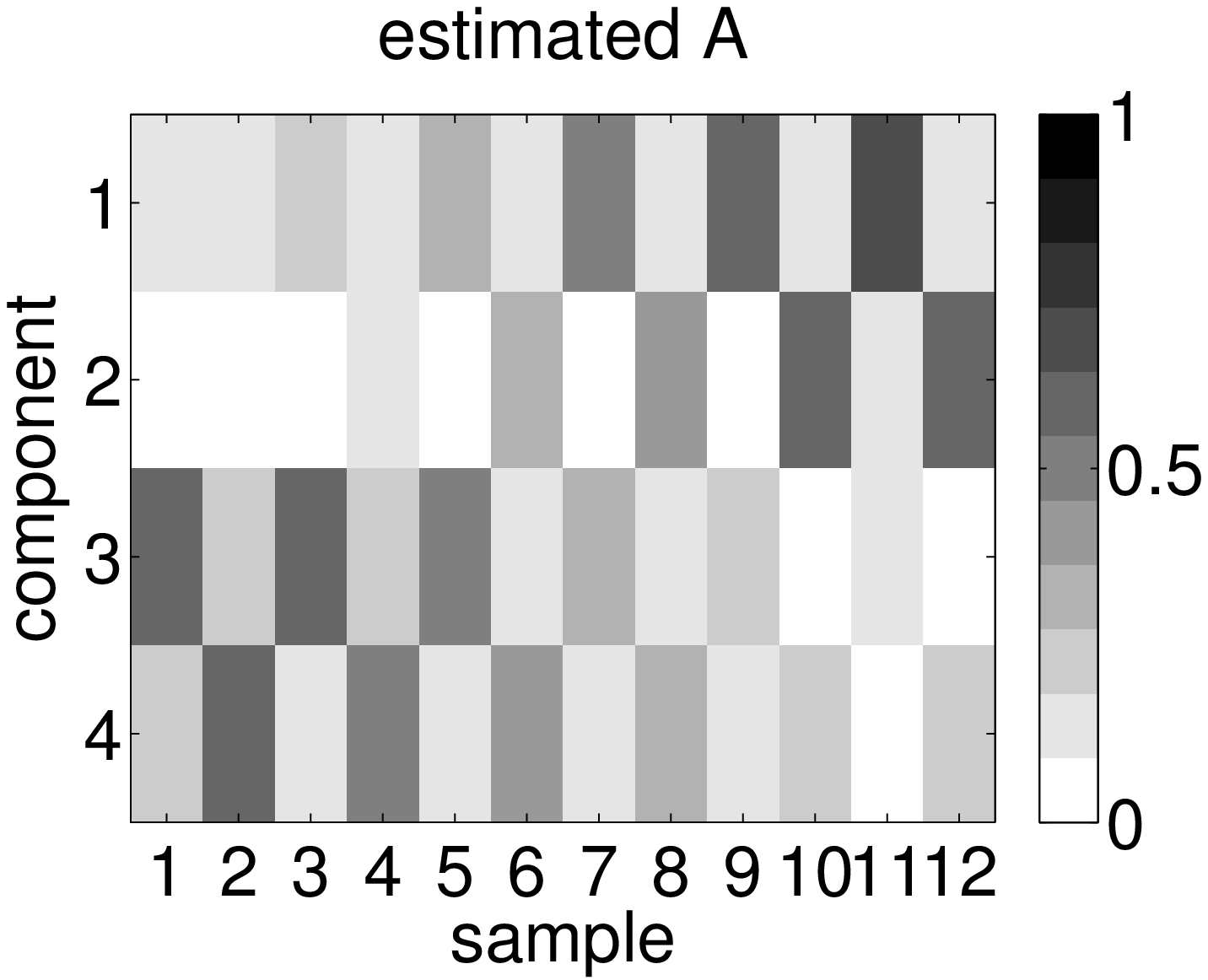}
&\hspace{-0.29cm} \includegraphics[height =0.16\textheight]{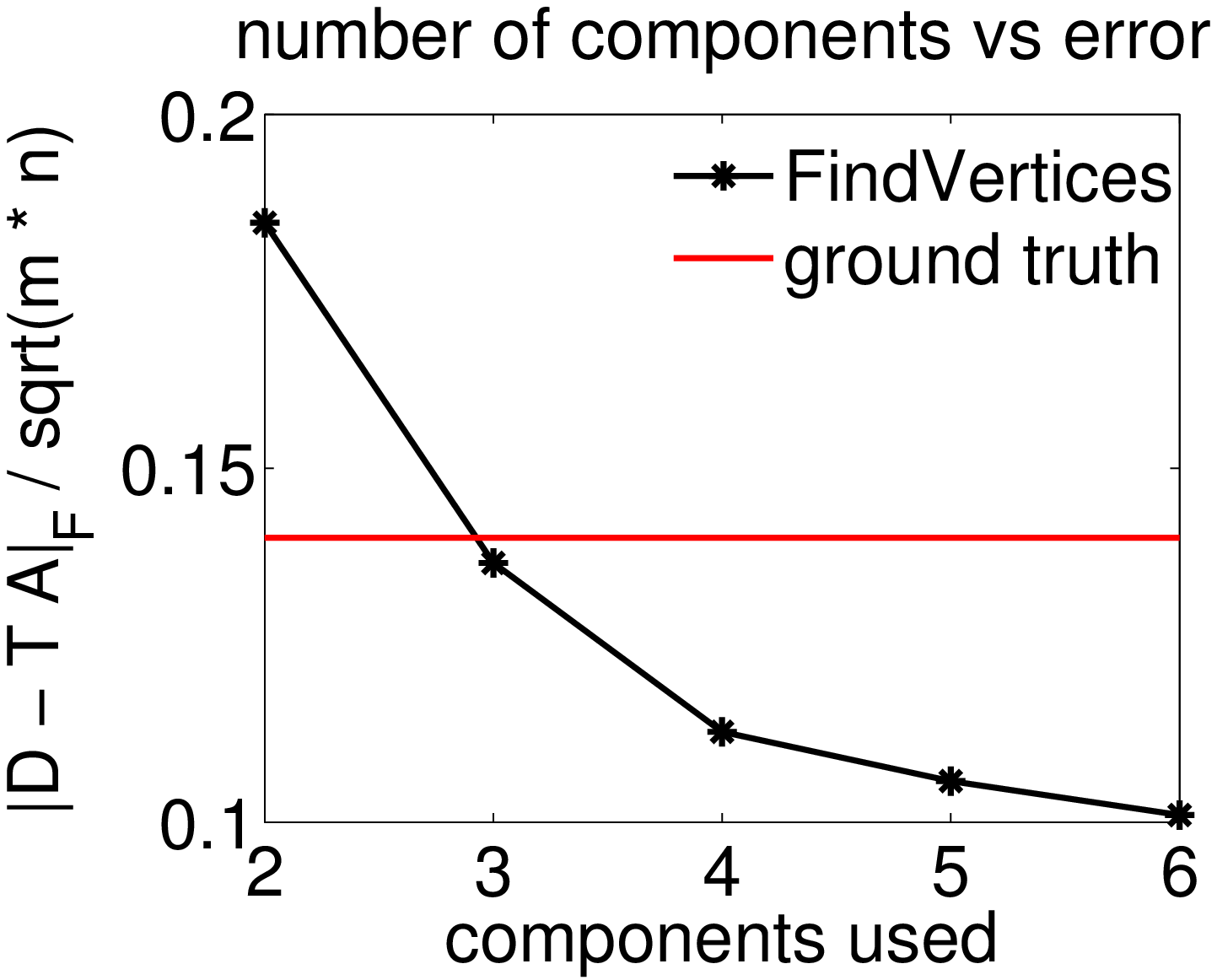}
\end{tabular}
\caption{Left: Mixture proportions of the ground truth. Middle: mixture
  proportions as estimated by our method. Right: RMSEs $\nnorm{D - \overline{T}
    \; \overline{A}}_F/(m \, n)^{1/2}$ in dependency of $r$.}\label{fig:real1}
\end{flushleft}
\end{figure}\hfill\\
\textbf{Analysis.} We apply our approach to obtain an approximate
factorization $D \approx \overline{T} \; \overline{A}$, $\overline{T} \in \{0,1\}^{m \times r}$,
$\overline{A} \in \R_+^{r \times n}$ and $\overline{A}^{\T} \bm{1}_r = \bm{1}_n$. We first
obtained $\overline{T}$ as outlined in Section \ref{sec:approximate}, replacing $\{0,1\}$ by
$\{0.1, 0.9\}$ in order to account for measurement noise in $D$ that slightly
pushes values towards 0.5. This can be accomodated re-scaling
$\wh{T}^{01}$ in step 4 of Algorithm \ref{alg:3} by $0.8$ and then adding $0.1$. Given $\overline{T}$, we solve the quadratic program   
$\overline{A} = \argmin_{A \in \R_+^{r \times n}, A^{\T} \bm{1}_r = \bm{1}_n}
\nnorm{D - \overline{T} A}_F^2$ and compare $\overline{A}$ to the ground truth $A^*$. In
order to judge the fit as well as the matrix $\overline{T}$ returned by our
method, we compute $T^* = \argmin_{T \in \{0,1\}^{m \times r}} \nnorm{D - T
  A^*}_F^2$ as in (9). We obtain 0.025 as average
mean squared difference of $\overline{T}$ and $T^*$, which corresponds to an agreement of 96
percent. Figure \ref{fig:real1} indicates at least a qualitative agreement of
$A^*$ and $\overline{A}$. In the rightmost plot, we compare the RMSEs of our
approach for different choices of $r$ relative to the RMSE of $(T^*,
A^*)$. The error curve flattens after $r = 4$, which suggests that with our
approach, we can recover the correct number of cell types.

\pagebreak

\appendix

\section{Proof of Proposition 1}

\renewcommand\thealgorithm{\arabic{algorithm}} 

Proposition 1 is about Algorithm \ref{alg:1}, which we re-state here.

\begin{algorithm}
\caption{\textsc{FindVertices} \textsc{exact}}\label{alg:1}
\algsetup{indent=0em}
\begin{algorithmic}
\STATE \begin{enumerate}\item[1.] Fix $p \in \aff(D)$ and compute $P = [D\cls{1} -p,\ldots,D\cls{n} - p]$.
     \end{enumerate}   
\STATE \begin{enumerate}\item[2.] Determine $r-1$ linearly independent columns
  $\mc{C}$ of $P$, obtaining $P\cls{\mc{C}}$ and subsequently $r-1$ linearly independent
  rows $\mc{R}$, obtaining $P_{\mc{R},\mc{C}} \in \R^{r-1 \times r-1}$.
\end{enumerate}
\STATE \begin{enumerate}\item[3.] Form $Z = P\cls{\mc{C}}
  (P_{\mc{R},\mc{C}})^{-1} \in \R^{m \times r-1}$ and $\wh{T} = Z (B^{(r-1)}  -
  p_{\mc{R}} \bm{1}_{2^{r-1}}^{\T}) + p \bm{1}_{2^{r-1}}^{\T} \in \R^{m \times
  2^{r-1}}$, where
the columns of $B^{(r-1)}$ correspond to the elements of $\{0,1\}^{r-1}$.\end{enumerate}
\STATE \begin{enumerate}\item[4.] Set $\mc{T} = \emptyset$. For $u =
  1,\ldots,2^{r-1}$, if $\wh{T}\cls{u} \in \{0,1 \}^m$ set $\mc{T} = \mc{T}
  \cup \{ \wh{T}\cls{u} \}$.
\end{enumerate}
\STATE  \begin{enumerate}\item[5.] Return $\mc{T} = \{0,1\}^m \cap \aff(D)$. \end{enumerate}
\end{algorithmic}
\end{algorithm}

\begin{prop}\label{prop:basic} The affine subspace $\text{aff}(D)$ contains no more than
  $2^{r-1}$ vertices of $[0,1]^m$. Moreover, Algorithm \ref{alg:1} provides
  all vertices contained in $\text{aff}(D)$.
\end{prop}

\begin{bew} Consider the first part of the statement. Let $b \in \{0,1\}^m$ and $p \in \aff(D)$ arbitrary. We have $b \in \aff(D)$
iff there exists $\theta \in \R^n$ s.t.  
\begin{equation}\label{eq:linsol_vert}
D \theta = b, \; \theta^{\T} \bm{1}_n = 1 \; \Longleftrightarrow \, \underbrace{[D\cls{1} -
p,\ldots,D\cls{n} - p]}_{=P} \theta + p = b \; \Longleftrightarrow \, P
\theta = b - p.
\end{equation}
Note that $\text{rank}(P) = r-1$. Hence, if there exists $\theta$
s.t. $P\theta = b - p$, such $\theta$ can be obtained from the unique $\lambda \in
\R^{r-1}$ solving $P_{\mc{R}, \mc{C}} \lambda = b_{\mc{R}} - p_{\mc{R}}$, where $\mc{R} \subset \{1,\ldots,m \}$ and $\mc{C}
\subset \{1,\ldots,n\}$ are subsets of rows respectively columns of $P$
s.t. $\text{rank}(P_{\mc{R}, \mc{C}}) = r-1$. Finally note that $b_{\mc{R}}
\in \{0,1\}^{r-1}$ so that there are no more than $2^{r-1}$ distinct right hand sides
$b_{\mc{R}} - p_{\mc{R}}$.\\ 
Turning to the second part of the statement, observe that for each $b \in \{0,1 \}^m$, there exists a unique $\lambda$
s.t. $P_{\mc{R},\mc{C}} \lambda = b_{\mc{R}} - p_{\mc{R}} \,\Leftrightarrow
\lambda = (P_{\mc{R},\mc{C}})^{-1} (b_{\mc{R}} - p_{\mc{R}})$. Repeating the argument
preceding \eqref{eq:linsol_vert}, if $b \in \{0,1\}^m \cap \aff(D)$, it must hold that
\begin{equation}\label{eq:lift_vert}
b = P\cls{\mc{C}} \lambda + p \,\Longleftrightarrow \, b =  \underbrace{P\cls{\mc{C}}
(P_{\mc{R}, \mc{C}})^{-1}}_{=Z} (b_{\mc{R}} - p_{\mc{R}}) + p
\,\Longleftrightarrow \, b = Z (b_{\mc{R}} - p_{\mc{R}}) + p.
\end{equation}
Algorithm \ref{alg:1} generates all possible right hand sides $\wh{T} = Z
(B^{(r-1)} - p_{\mc{R}} \bm{1}_{2^{r-1}}^{\T}) + p \bm{1}_{2^{r-1}}^{\T}$, where
$B^{(r-1)}$ contains all elements of $\{0,1\}^{r-1}$ as its
columns. Consequently if $b \in \{0,1\}^{m} \cap \aff(D)$, it must
appear as a column of $\wh{T}$. Conversely, if the leftmost equality in
\eqref{eq:lift_vert} does not hold, $b \notin \aff(D)$ and the 
column of $\wh{T}$ corresponding to $b_{\mc{R}}$ cannot be a binary vector. 
\end{bew}

\renewcommand\thealgorithm{\thesection.\arabic{algorithm}} 

\section{The matrix factorization problem without the constraint $A^{\T}
\mathbf{1}_r = \mathbf{1}_n$}

In the paper, we have provided Algorithm 2 to solve the matrix factorization problem 
\begin{equation}
\text{find} \; T \in \{0,1\}^{m \times r} \; \, \text{and} \; \, A \in \R^{r
  \times n}, \, A^{\T} \bm{1}_r = \bm{1}_n \; \; \text{such that} \; \, D = TA.
\end{equation}
We here provide variants of Algorithms 1 and 2 to solve the corresponding
problem without the constraint $A^{\T} \bm{1}_r = \bm{1}_n$, that is   
\begin{equation}\label{eq:exactfactorize_linear}
\text{find} \; T \in \{0,1\}^{m \times r} \; \, \text{and} \; \, A \in \R^{r
  \times n}  \; \, \text{such that} \; \, D = TA.
\end{equation}
The following Algorithm \ref{alg:3} is the analog of Algorithm 1. Algorithm
\ref{alg:3} yields $\text{span}(D) \cap \{0,1 \}^m$, which can be proved along
the lines of the proof of Proposition 1 under the stronger assumption that $T$ has $r$
\emph{linearly} independent in place of only $r$ \emph{affinely} independent
columns, which together with the assumption $\text{rank}(A) = r$ implies that
also $\text{rank}(D) = r$ (cf.~Section 2.1 of the paper). Algorithm \ref{alg:3} results from Algorithm
1 by setting $p = 0$ and replacing $r-1$ by $r$. 
\renewcommand{\thesection}{\Alph{section}} 
\vspace{-0.2cm}
\begin{center}
\begin{algorithm}
\caption{\textsc{FindVertices} \textsc{exact\_linear}}\label{alg:3}
\algsetup{indent=0em}
\begin{algorithmic}
\STATE \begin{enumerate}\item[1.] Determine $r$ linearly independent columns
  $\mc{C}$ of $D$, obtaining $D\cls{\mc{C}}$ and subsequently $r$ linearly independent
  rows $\mc{R}$, obtaining $D_{\mc{R},\mc{C}} \in \R^{r \times r}$.
\end{enumerate}
\STATE \begin{enumerate}\item[2.] Form $Z = D\cls{\mc{C}}
  (D_{\mc{R},\mc{C}})^{-1} \in \R^{m \times r}$ and $\wh{T} = Z B^{(r)}  \in \R^{m \times
  2^{r}}$, where
the columns of $B^{(r)}$ correspond to the elements of $\{0,1\}^{r}$
\end{enumerate}
\STATE \begin{enumerate}\item[3.] Set $\mc{T} = \emptyset$. For $u =
  1,\ldots,2^{r}$, if $\wh{T}\cls{u} \in \{0,1 \}^m$ set $\mc{T} = \mc{T}
  \cup \{ \wh{T}\cls{u} \}$.
\end{enumerate}
\STATE  \begin{enumerate}\item[4.] Return $\mc{T} = \{0,1\}^m \cap \text{span}(D)$. \end{enumerate}
\end{algorithmic}
\end{algorithm} 
\end{center}
The following Algorithm \ref{alg:4} solves problem
\eqref{eq:exactfactorize_linear} given the output of Algorithm \ref{alg:3}.
\begin{algorithm}
\caption{\textsc{BinaryFactorization} \textsc{exact\_linear}}\label{alg:4}
\begin{algorithmic}
\STATE \begin{enumerate}\item[1.] Obtain $\mc{T}$ as output from \textsc{FindVertices Exact\_linear}($D$)
     \end{enumerate}   
\STATE \begin{enumerate}\item[2.] Select $r$ linearly independent elements of
  $\mc{T}$ to be used as columns of $T$. 
\end{enumerate}
\STATE \begin{enumerate}\item[3.] Obtain $A$ as solution of the linear system
  $ T A = D$. 
\end{enumerate}
\STATE  \begin{enumerate}\item[4.] Return $(T,A)$ solving problem \eqref{eq:exactfactorize_linear}. \end{enumerate}
\end{algorithmic}
\end{algorithm} \hfill \\
For the sake of completness, we provide Algorithm \ref{alg:5} as a counterpart
to Algorithm 3 regarding the approximate case. An additional modification is
necessary to eliminate the zero vector, which is always contained in
$\text{span}(D)$ and hence would be returned as a column of 
$T$ if we used $B^{(r)}$ in place of $B_{\setminus 0}^{(r)}$ in step 2. below,
whose columns correspond to the elements of $\{0,1\}^r \setminus \{0_r \}$. 
\vspace{-0.2cm}
\begin{algorithm}
\caption{\textsc{FindVertices} \textsc{approximate\_linear}}\label{alg:5}
\begin{algorithmic}
\STATE \begin{enumerate}\item[1.] Compute $U^{(r)}  \in \R^{m
    \times r}$, the left singular vectors corresponding to the $r$ largest singular values of
  $D$. Select $r$ linearly independent rows $\mc{R}$ of $U^{(r)}$,
  obtaining $U_{\mc{R},:}^{(r)} \in \R^{r \times r}$. 
\end{enumerate}
\STATE \begin{enumerate}\item[2.] Form $Z =  U^{(r)}
  (U_{\mc{R},:}^{(r)})^{-1}$ and $\wh{T} = Z B_{\setminus 0}^{(r)}$. \end{enumerate}
\STATE \begin{enumerate}\item[4.] Compute $\wh{T}^{01} \in \R^{m \times
    2^{r}}$: for $u =  1,\ldots,2^{r}$, $i =
  1,\ldots,m$, set $\wh{T}_{i,u}^{01} = I(\wh{T}_{i,u} > \frac{1}{2})$.
\end{enumerate}
\STATE  \begin{enumerate}\item[5.] For $u=1,\ldots,2^{r}$, set $\delta_u =
  \nnorm{\wh{T}\cls{u} - \wh{T}\cls{u}^{01}}_2$. Order increasingly s.t. $\delta_{u_1}
  \leq \ldots \leq \delta_{2^{r}}$.\end{enumerate}
\STATE \begin{enumerate}\item[6.] Return $T = [\wh{T}\cls{u_1}^{01} \ldots \wh{T}\cls{u_r}^{01}]$ 
\end{enumerate}
\end{algorithmic}
\end{algorithm}

\section{Matrix factorization with left and right binary factor and
  real-valued middle factor}

We here sketch how our approach can be applied to obtain a matrix
factorization considered in \cite{Meeds2007}, which is of the form $T W A^{\T}$ with both $T$ and $A$ binary
and $W$ real-valued in the exact case; the noisy case be tackled
similarly with the help of Algorithm \ref{alg:5} and is thus omitted.\\ Consider
the matrix factorization problem
\begin{equation}\label{eq:exactfactorize_threeway}
\text{find} \; T \in \{0,1\}^{m \times r}, \; \, A \in \{0,1\}^{n \times
  r}  \; \, \text{and} \; \,  W \in \R^{r \times r}\; \, \text{such
  that} \; \, D = T W A^{\T},
\end{equation}
and suppose that $\text{rank}(D) = r$. Then the following Algorithm
\ref{alg:6} solves problem \eqref{eq:exactfactorize_threeway}.
\begin{algorithm}[t!]
\caption{\textsc{ThreeWayBinaryFactorization}}\label{alg:6}
\begin{algorithmic}
\STATE \begin{enumerate}\item[1.] Obtain $\mc{T}$ as output from \textsc{FindVertices Exact\_linear}($D$)
     \end{enumerate}   
\STATE \begin{enumerate}\item[2.] Obtain $\mc{A}$ as output from \textsc{FindVertices Exact\_linear}($D^{\T}$)
     \end{enumerate}  
\STATE \begin{enumerate}\item[3.] Select $r$ linearly independent elements of
  $\mc{T}$ and $\mc{A}$ to be used as columns of $T$ respectively $A$. 
\end{enumerate}
\STATE \begin{enumerate}\item[4.] Obtain $W = (T^{\T} T)^{-1} T^{\T} D A
  (A^{\T} A)^{-1}$. 
\end{enumerate}
\STATE  \begin{enumerate}\item[5.] Return $(T,A,W)$ solving problem \eqref{eq:exactfactorize_threeway}. \end{enumerate}
\end{algorithmic}
\end{algorithm} \hfill \\

\section{Proof of Corollary 1}
Corollary 1 follows directly from Proposition 1. 

\section{Proof of Proposition 2}
 
\setcounter{propo}{1}

\renewcommand{\thesection}{\arabic{section}}

Before re-stating Proposition \ref{prop:separability} below, let 
us recall problem (1) and property (4) of the paper. 
\begin{align*}
&\text{find} \; T \in \{0,1\}^{m \times r} \; \, \text{and} \; \, A \in \R^{r
  \times n}, \, A^{\T} \bm{1}_r = \bm{1}_n \; \; \text{such that} \; \, D =
TA. \;(1)\\
\\
&\aff(D) \cap \{0,1\}^m = \aff(T) \cap \{0,1\}^m = \{T_{:,1},\ldots,T_{:,r} \}
\;(4)
\end{align*}

Let us also recall that $T$ is said to be \emph{separable} if there exists a permutation
$\Pi$ such that $\Pi T = [M; I_r]$, where $M \in \{0,1 \}^{m-r \times r}$.

\begin{prop}\label{prop:separability} If $T$ is separable, condition (4) holds
and thus problem (1) has a
unique solution.  
\end{prop}
\begin{bew} We have $\aff(T) \ni b \in \{0,1\}^m$ iff there exists $\lambda \in \R^r, \, \lambda^{\T}
 \bm{1}_r = 1$ such that 
\begin{equation*}
T \lambda = b \; \Longleftrightarrow \; \Pi T \lambda = \Pi b \;
\Longleftrightarrow \;  [M; I_r] \lambda = \Pi b. 
\end{equation*}
Since $\Pi b \in \{0,1\}^m$, for the bottom $r$ block of the linear system to
be fulfilled, it is necessary that $\lambda \in \{0,1\}^r$. The condition
$\lambda^{\T} \bm{1}_r = 1$ then implies that $\lambda$ must be one of the $r$
canonical basis vectors of $\R^r$. We conclude that $\aff(T) \cap \{0,1\}^m =
\{T_{:,1},\ldots,T_{:,r} \}$.  
\end{bew}

\renewcommand{\thesection}{\Alph{section}} 

\section{Proof of Theorem 1}
Our proof of Theorem 1 relies on two seminal results on random
$\pm1$-matrices.
\begin{theoApp} \label{theo:kahn}\cite{Kahn1995} Let $M$ be a random $m \times r$-matrix whose entries are drawn
  i.i.d.~from $\{-1,1\}$ each with probability $\frac{1}{2}$. There is a constant
  $C$ so that if $r \leq m-C$,
\begin{equation}\label{eq:prob_kahn}
\p \left(\text{span}(M) \cap \{-1,1\}^m = \{\pm M_{:,1},\ldots, \pm M_{:,r} \} \right) \geq 1 - (1
+ o(1)) \, 4 \binom{r}{3} \left( \frac{3}{4} \right)^m \; \, \text{as} \; m
\rightarrow \infty. 
\end{equation}
\end{theoApp} 
\begin{theoApp}\cite{Tao2007} Let $M$ be a random $m \times r$-matrix, $r \leq
  m$, whose entries are drawn
  i.i.d.~from $\{-1,1\}$ each with probability $\frac{1}{2}$. Then 
\begin{equation}\label{eq:prob_tao}
\p \big(\text{M} \, \text{has linearly independent columns} \big) \geq 1 -
\left(\frac{3}{4} + o(1) \right)^m  \; \, \text{as} \; m
\rightarrow \infty.  
\end{equation}
\end{theoApp}
We are now in position to re-state and prove Theorem 1.
\begin{theo} \label{corro:kahn} Let $T$ be a random $m \times r$-matrix 
whose entries are drawn i.i.d.~from $\{0,1\}$ each with probability
$\frac{1}{2}$. Then, there is a constant
  $C$ so that if $r \leq m-C$,
\begin{equation*} 
\p \Big( \aff(T) \cap \{0,1\}^m =
\{T_{:,1},\ldots,T_{:,r} \Big) \geq 1 - (1
+ o(1)) \, 4 \binom{r}{3} \left( \frac{3}{4} \right)^m - \left( \frac{3}{4} +
  o(1) \right)^m\; \, \text{as} \; m
\rightarrow \infty. 
\end{equation*}
\end{theo}
\begin{bew} Note that $T = \frac{1}{2} (M + \bm{1}_{m \times r})$, where
$M$ is a random $\pm1$-matrix as in Theorem \ref{theo:kahn}. Let $\lambda \in
\R^{r}$, $\lambda^{\T} \bm{1}_r  = 1$  and $b \in \{0,1\}^m$. Then 
\begin{equation}\label{eq:corro_kahn_proof}
T \lambda = b \; \Longleftrightarrow \; \frac{1}{2} (M \lambda  + \bm{1}_m) = b
\; \Longleftrightarrow \, M \lambda = 2 b - \bm{1}_m \in \{-1,1\}^m. 
\end{equation} 
Now note that with the probability given in \eqref{eq:prob_kahn}, 
\begin{equation*} 
\text{span}(M) \cap \{-1,1\}^m = \{\pm M_{:,1},\ldots, \pm M_{:,r} \} \; \Longrightarrow \, \text{aff}(M) \cap
\{-1,1\}^{m} \subseteq \{\pm M_{:,1},\ldots,\pm M_{:,r} \}
\end{equation*}
On the other hand, with the probability given in \eqref{eq:prob_tao}, the
columns of $M$ are linearly independent. If this is the case, 
\begin{align}\label{eq:desired}
&\qquad \;\,\text{aff}(M) \cap
\{-1,1\}^{m} \subseteq \{\pm M_{:,1},\ldots,\pm M_{:,r} \}  \notag \\
& \Longrightarrow \, \text{aff}(M) \cap
\{-1,1\}^{m} = \{M_{:,1},\ldots,M_{:,r} \}. 
\end{align}
To verify this, first note the obvious inclusion 
$\text{aff}(M) \cap
\{-1,1\}^{m} \supseteq \{M_{:,1},\ldots,M_{:,r} \}$. Moreover, suppose by contradiction that there exists $j \in \{1,\ldots,r\}$
and $\theta \in \R^{r}$, $\theta^{\T} \bm{1}_r = 1$ such that $M \theta =
-M_{:,j}$. Writing $e_j$ for the $j$-th canonical basis vector, this would
imply $M (\theta + e_j) = 0$ and in turn by linear independence $\theta = -e_j$, which contradicts
$\theta^{\T} \bm{1}_r = 1$.\\ 
Under the event \eqref{eq:desired}, $M \lambda = 2 b - \bm{1}_m$ is fulfilled iff
$\lambda$ is equal to one of the canonical basis vectors and $2 b - \bm{1}_m$ equals the
corresponding column of $M$. We conclude the assertion in view of
\eqref{eq:corro_kahn_proof}.  
\end{bew}

\section{Theorem 1: empirical evidence}

\renewcommand\thefigure{\thesection.\arabic{figure}}    

\begin{figure}[h!] 
\begin{center}

\includegraphics[height = 0.24\textheight]{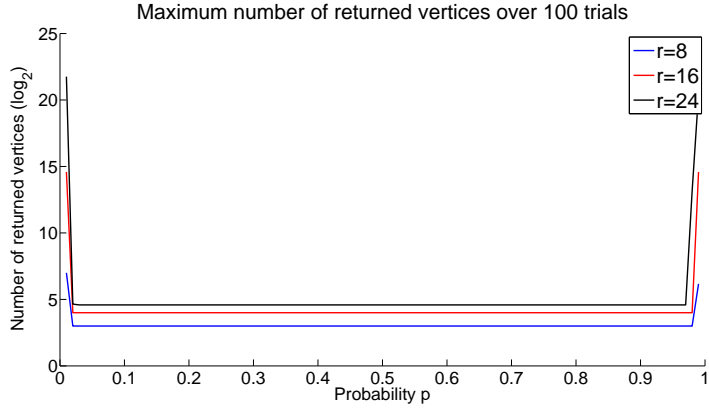}

\caption{Number of vertices contained in $\aff(T)$ over 100 trials for $T$ drawn entry-wise from a Bernoulli
distribution with parameter $p$.}\label{fig:uniqueness}
\end{center}
\end{figure}

It is natural to ask whether a result similar to Theorem \ref{corro:kahn} holds
if the entries of $T$ are drawn from a Bernoulli distribution with parameter 
$p$ in $(0,1)$ sufficiently far away from the boundary points. We have conducted
an experiment whose outcome suggests that the answer is positive. For this
experiment, we consider the grid $\{0.01,0.02,\ldots,0.99\}$ for $p$ and
generate random binary matrices $T \in \R^{m \times r}$ with $m = 500$ and
$r \in \{8,16,24\}$ whose entries are i.i.d.~Bernoulli with parameter $p$. 
For each value of $p$ and $r$, 100 trials are considered, and for each of
these trials, we compute the number of vertices of $[0,1]^m$ contained
in $\text{aff}(T)$. In Figure \ref{fig:uniqueness}, we report the maximum
number of vertices over these trials. One observes that except for a small
set of values of $p$ very close to $0$ or $1$, exactly $r$ vertices are
returned in all trials. On the other hand, for extreme values of $p$
the number of vertices can be as large as $2^{20}$ in the worst case.    

\section{Entire set of experiments with synthetic data}

\setcounter{figure}{0}

In section 4.1 of the paper, we have presented only a subset of all synthetic
data experiments that we have performed. We here present the entire set.\\
For the first set of experiments, we have considered three different
setups concerning the generation of $T$ and $A$ and two choices of $r$ (10 and
20),  out of which only the results of the first one ('T0.5') for $r = 10$ are
reported in the paper.\\
\textbf{Setups.}\\
\emph{'T0.5':} We generate $D = T^* A^* + \alpha E$, where the entries of
$T^*$ are drawn i.i.d.~from $\{0,1\}$ with probability 0.5, the columns of $A$
are drawn i.i.d.~uniformly from the probability simplex and the entries of $E$
are i.i.d.~standard Gaussian. We let $m = 1000$, $r \in \{10,20 \}$, $n =
2r$, and let the noise level $\alpha$ vary along a grid starting from 0.\\
\emph{'Tsparse+dense':} The matrix $T$ is now generated by drawing the
entries of one half of the columns of $T$ i.i.d.~from a Bernoulli distribution
with probability $0.1$ ('sparse' part), and the second half from a Bernoulli
distribution with parameter $0.9$ ('dense' part). The rest is as for the first
setup.  
\emph{'T0.5,Adense'}: As for \emph{'T0.5'} apart from the following
modification: after random generation of $A$ as above, we compute its
Euclidean projection on $\{A \in \R_{+}^{r \times n}:\,A^{\T}\bm{1}_r =
\bm{1}_n, \; \max_{k,i} A_{k,i} \leq 2/r \}$, thereby constraining the columns
of $A$ to be roughly constant. With such $A$, all data points are situated
near the barycentre $T \bm{1}_r/r$ of the simplex generated by the columns of
$T$. Given that the goal is to recover vertices, this setup is hence potentially more
difficult.
\begin{figure}[h!]
\begin{flushleft}
\begin{tabular}{lll}
\hline
\hspace{-0.28cm}\includegraphics[height = 0.15\textheight]{T0.5_err_m_1000_k_10.eps}
&\hspace{-0.28cm} \includegraphics[height = 0.15\textheight]{T0.5_denoised_m_1000_k_10.eps}
&\hspace{-0.29cm} \includegraphics[height =
0.15\textheight]{T0.5_objective_m_1000_k_10.eps}\\
\hspace{-0.28cm}\includegraphics[height = 0.15\textheight]{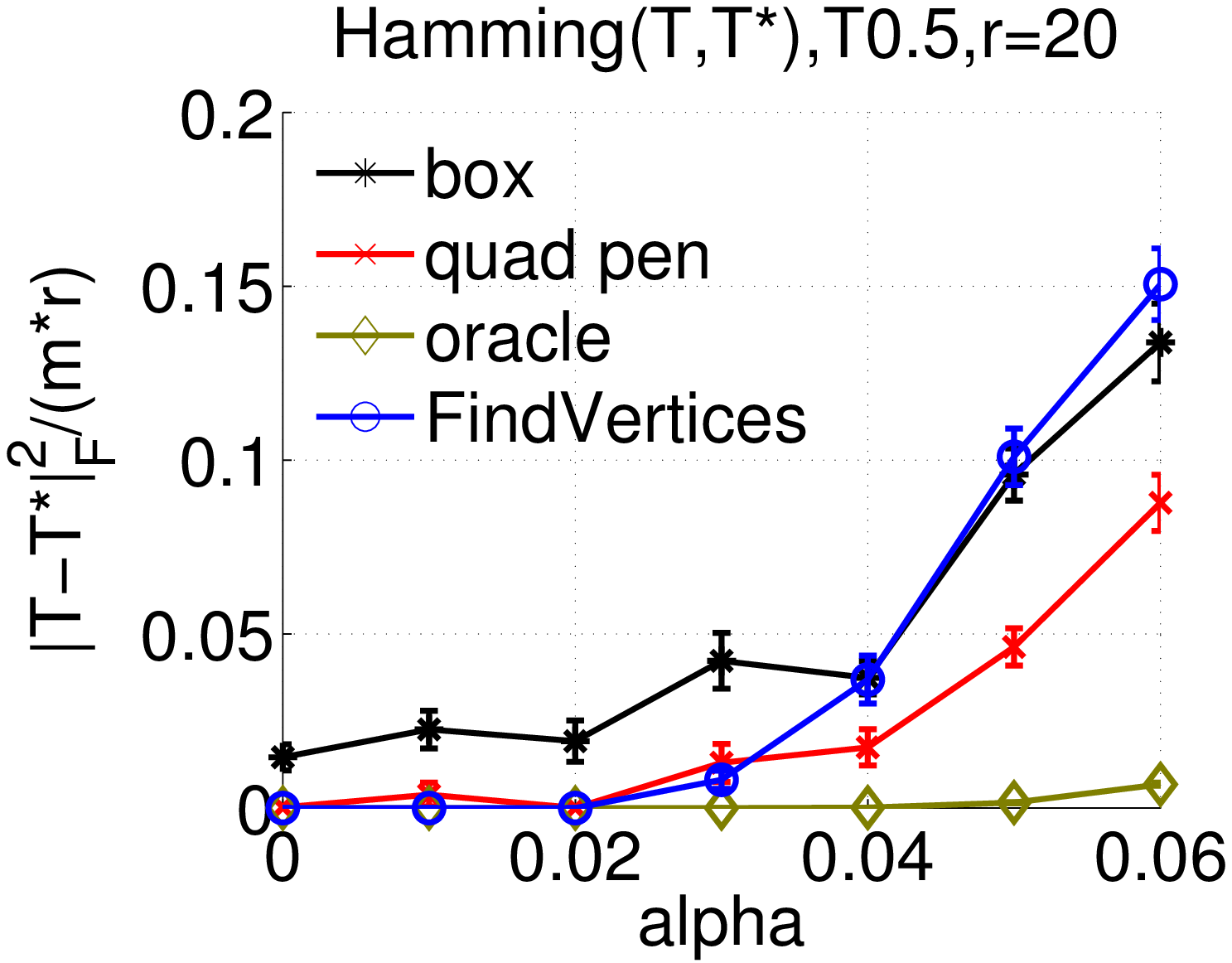}
& \hspace{-0.28cm}\includegraphics[height = 0.15\textheight]{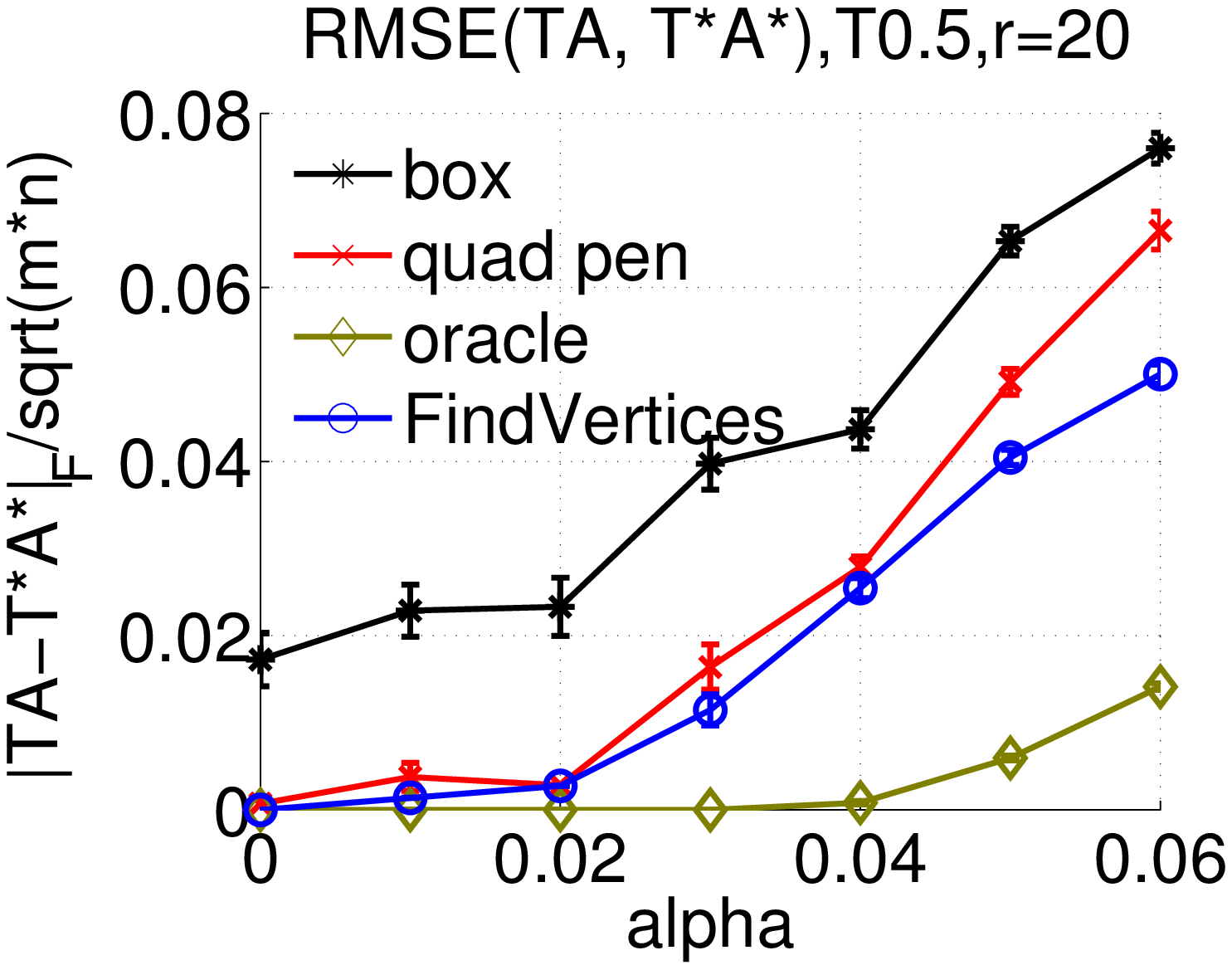}
& \hspace{-0.29cm}\includegraphics[height =
0.15\textheight]{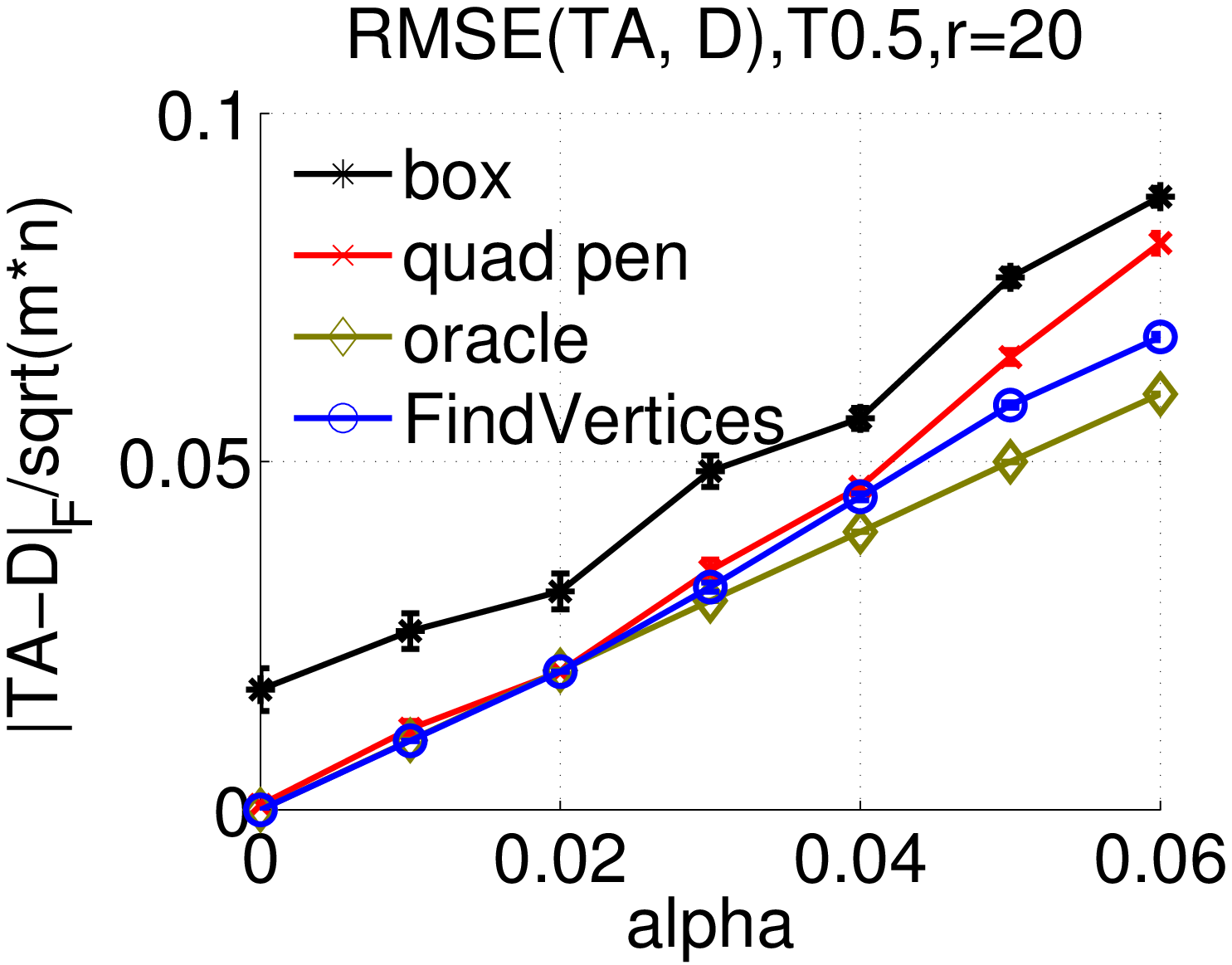}\\
\hline
\hspace{-0.28cm}\includegraphics[height = 0.15\textheight]{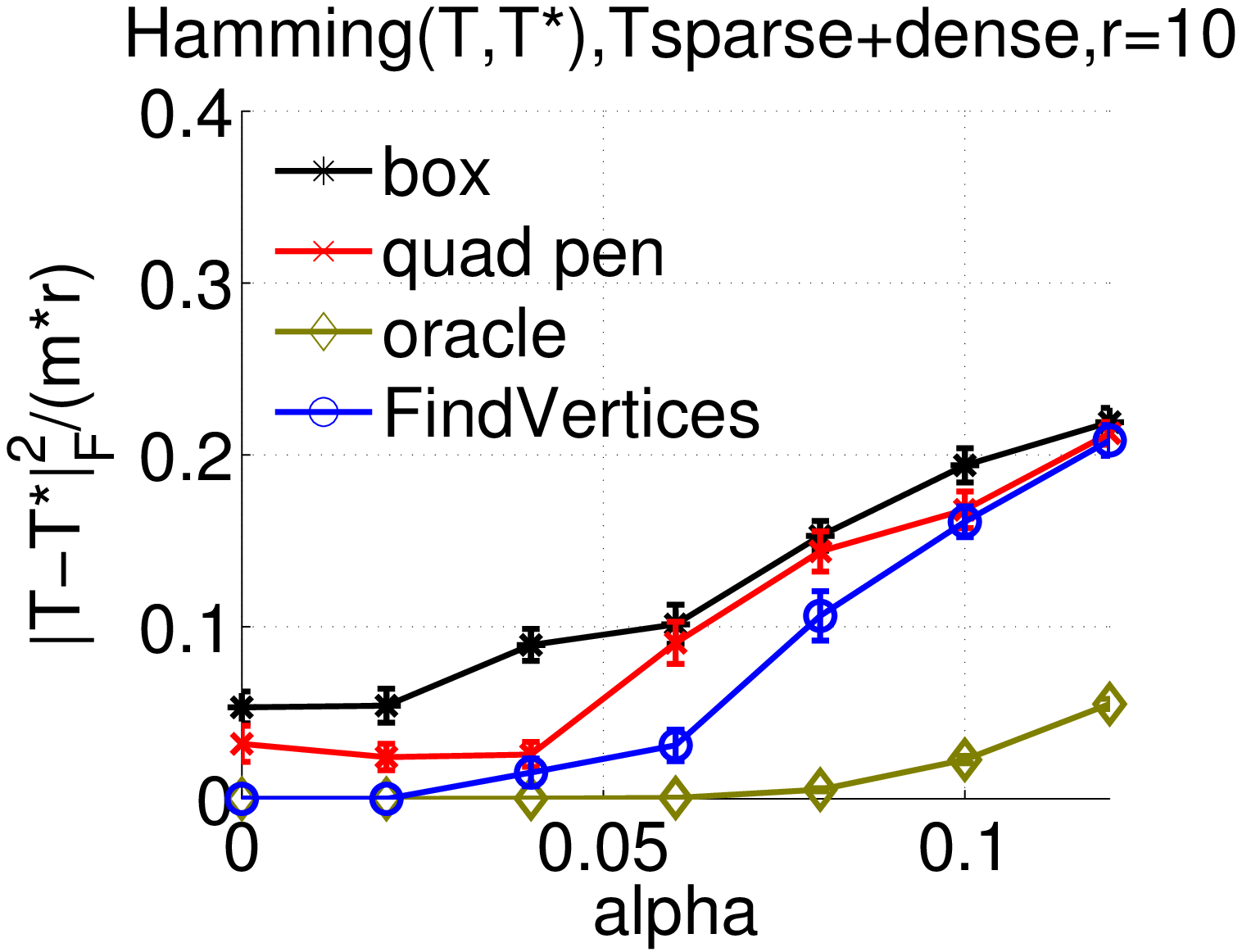}
&\hspace{-0.28cm} \includegraphics[height = 0.15\textheight]{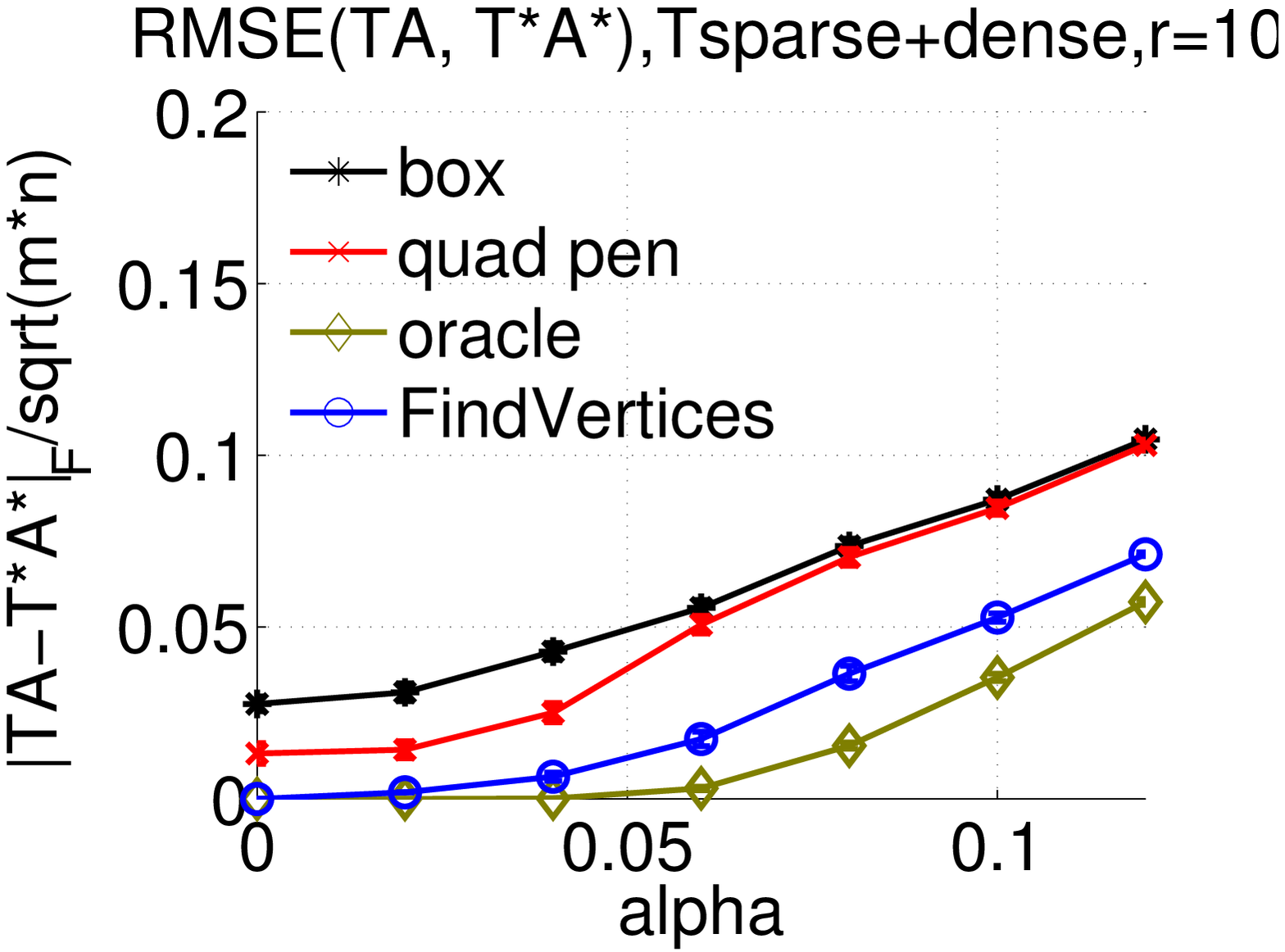}
&\hspace{-0.29cm} \includegraphics[height =
0.15\textheight]{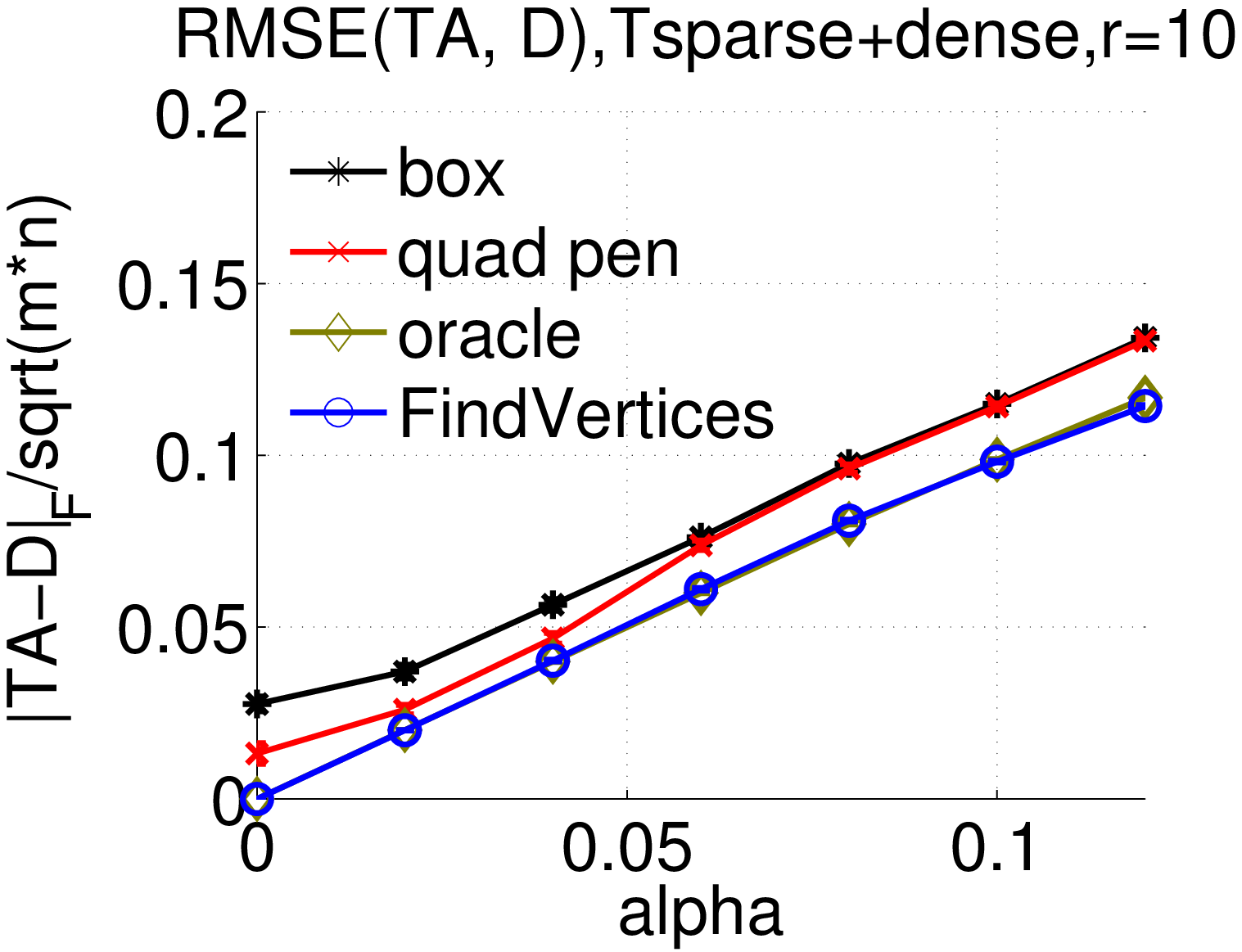}\\
\hspace{-0.28cm}\includegraphics[height = 0.15\textheight]{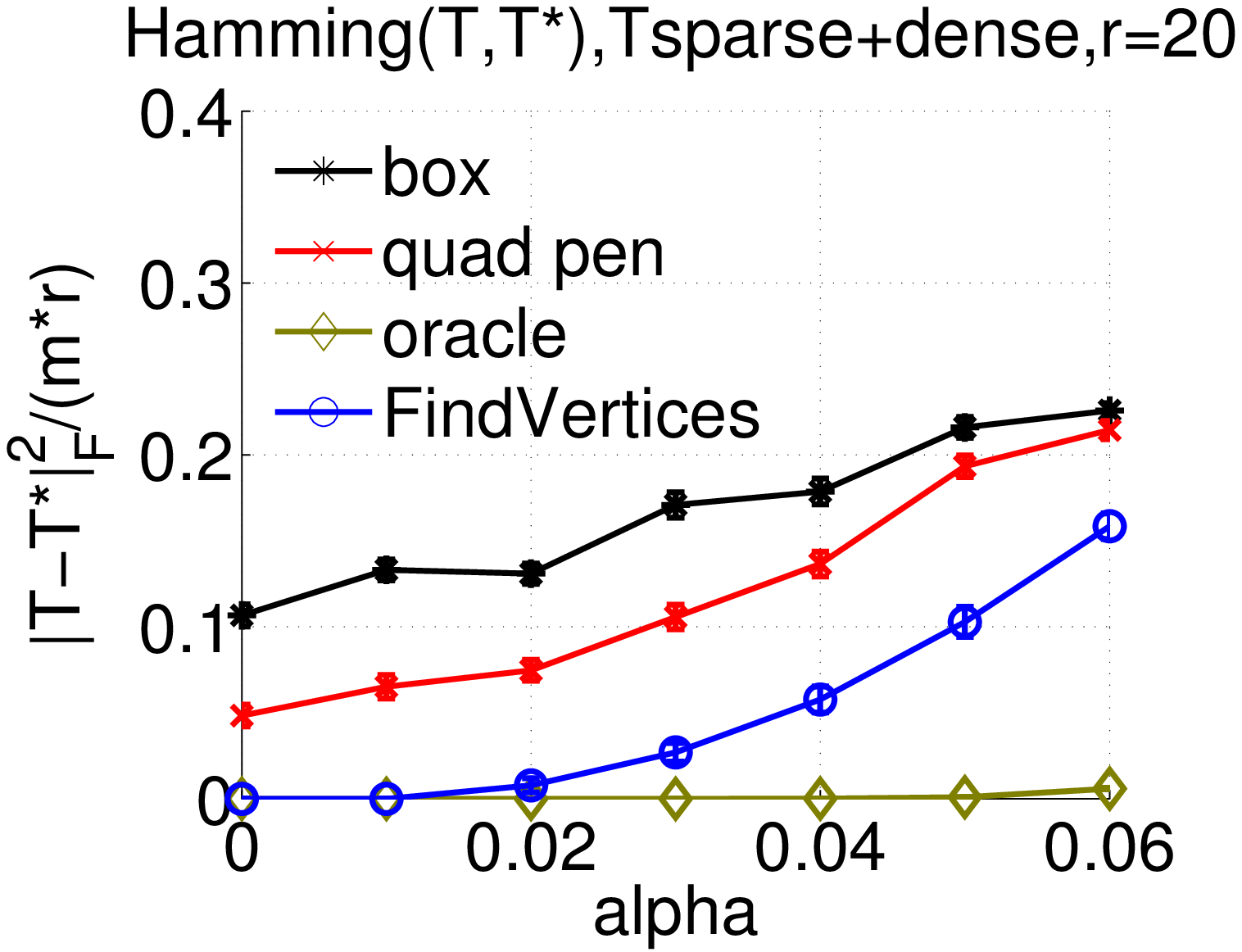}
& \hspace{-0.28cm}\includegraphics[height = 0.15\textheight]{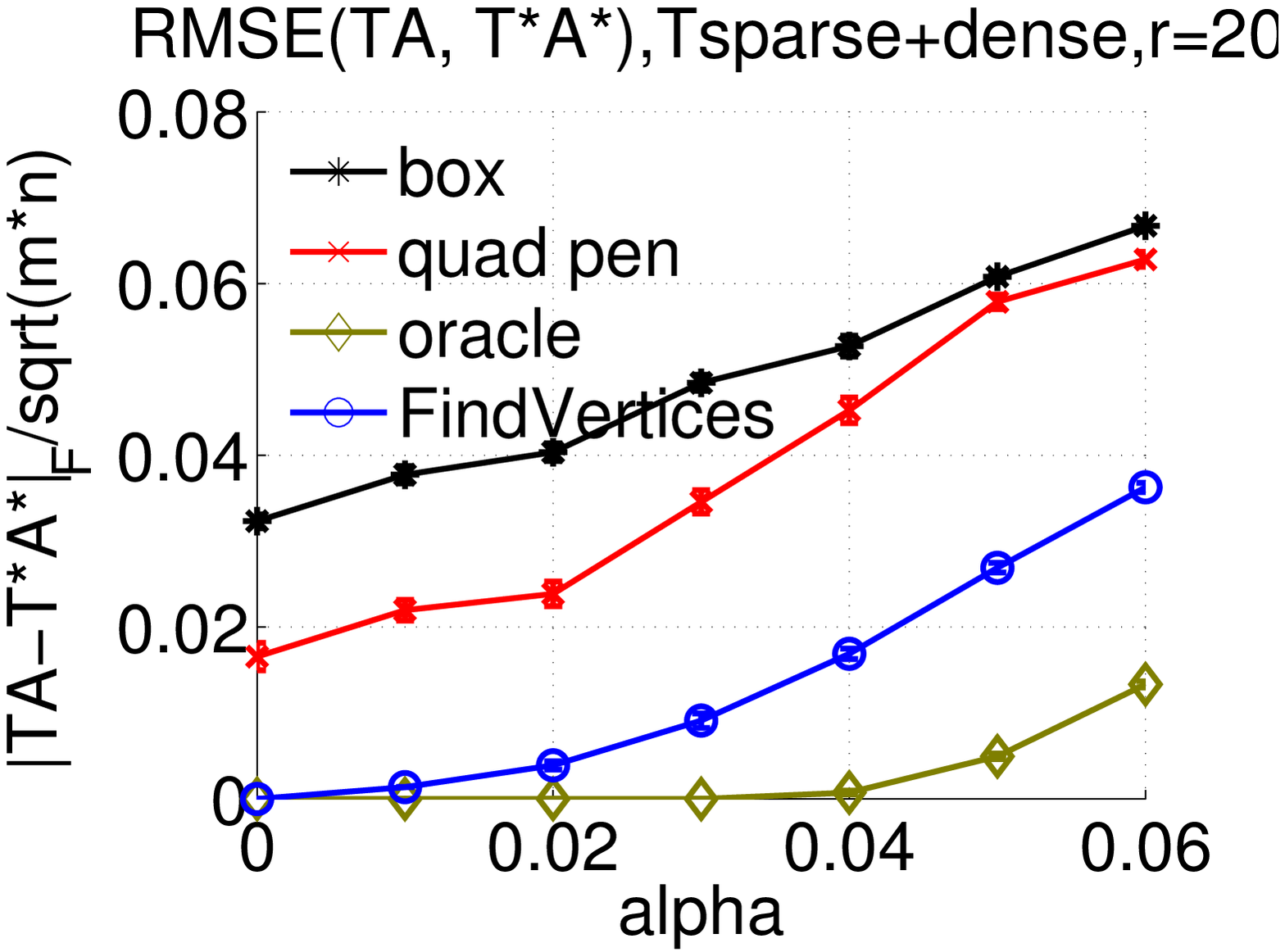}
& \hspace{-0.29cm}\includegraphics[height =
0.15\textheight]{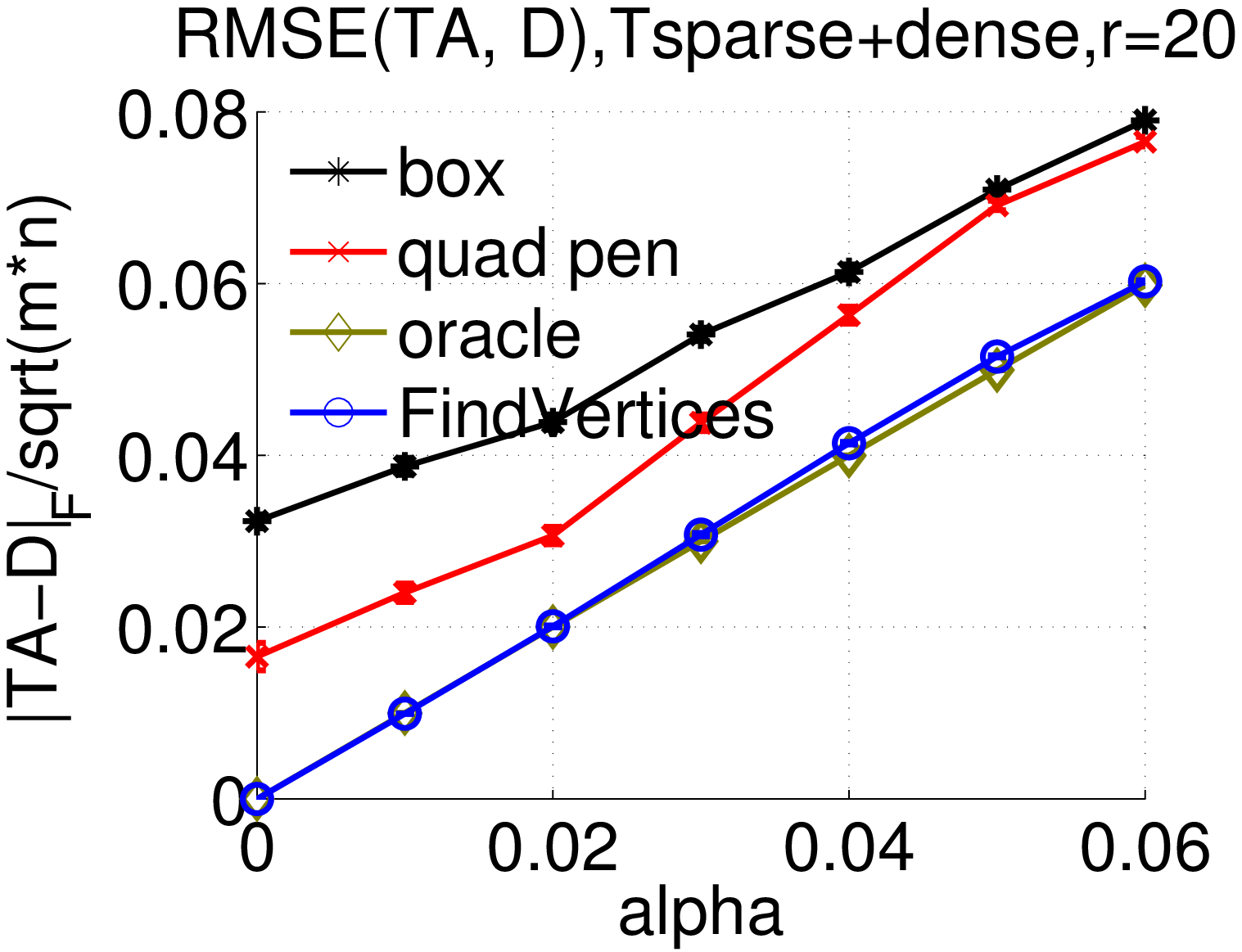}\\
\hline
\hspace{-0.28cm}\includegraphics[height =
0.15\textheight]{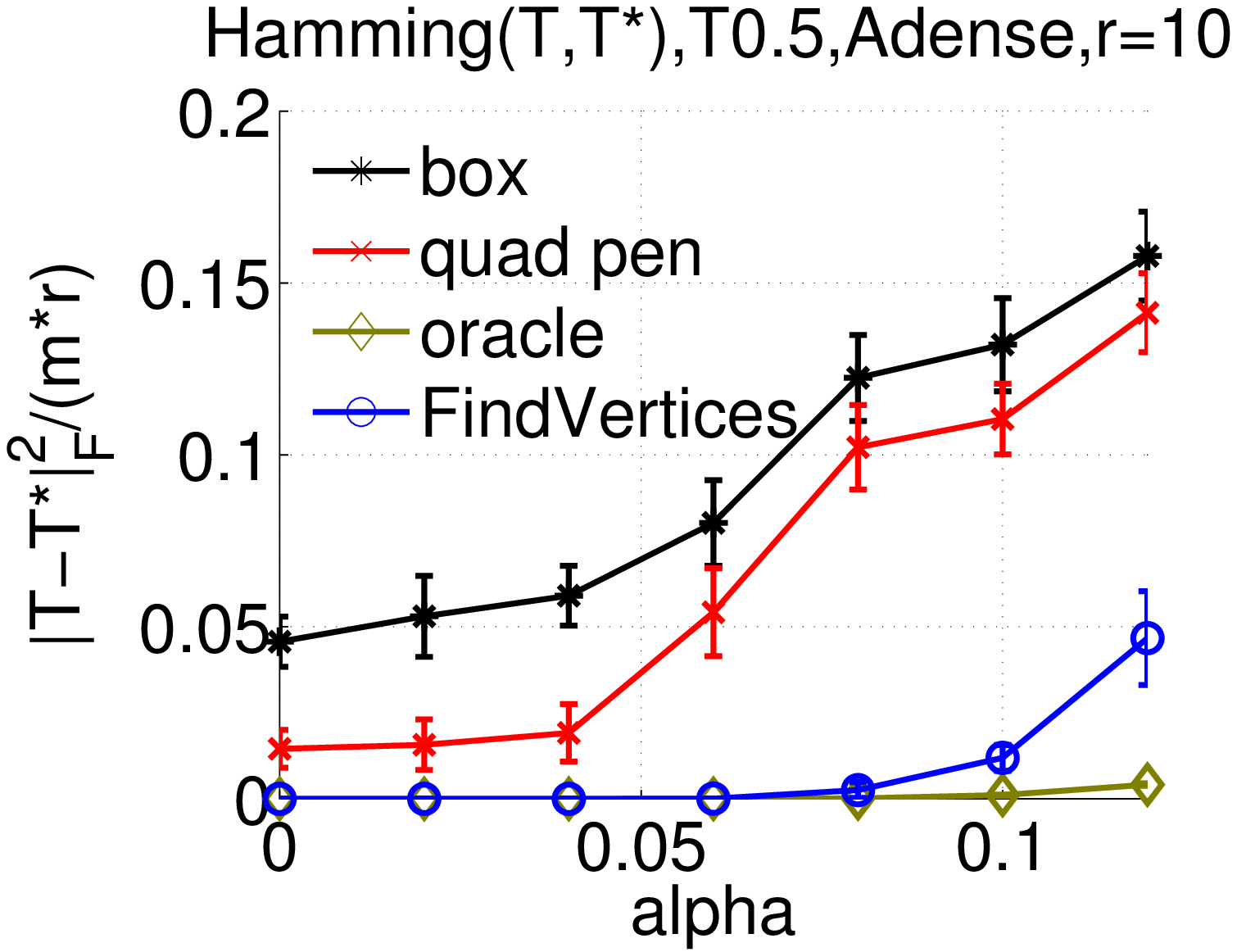}
 &\hspace{-0.28cm} \includegraphics[height = 0.15\textheight]{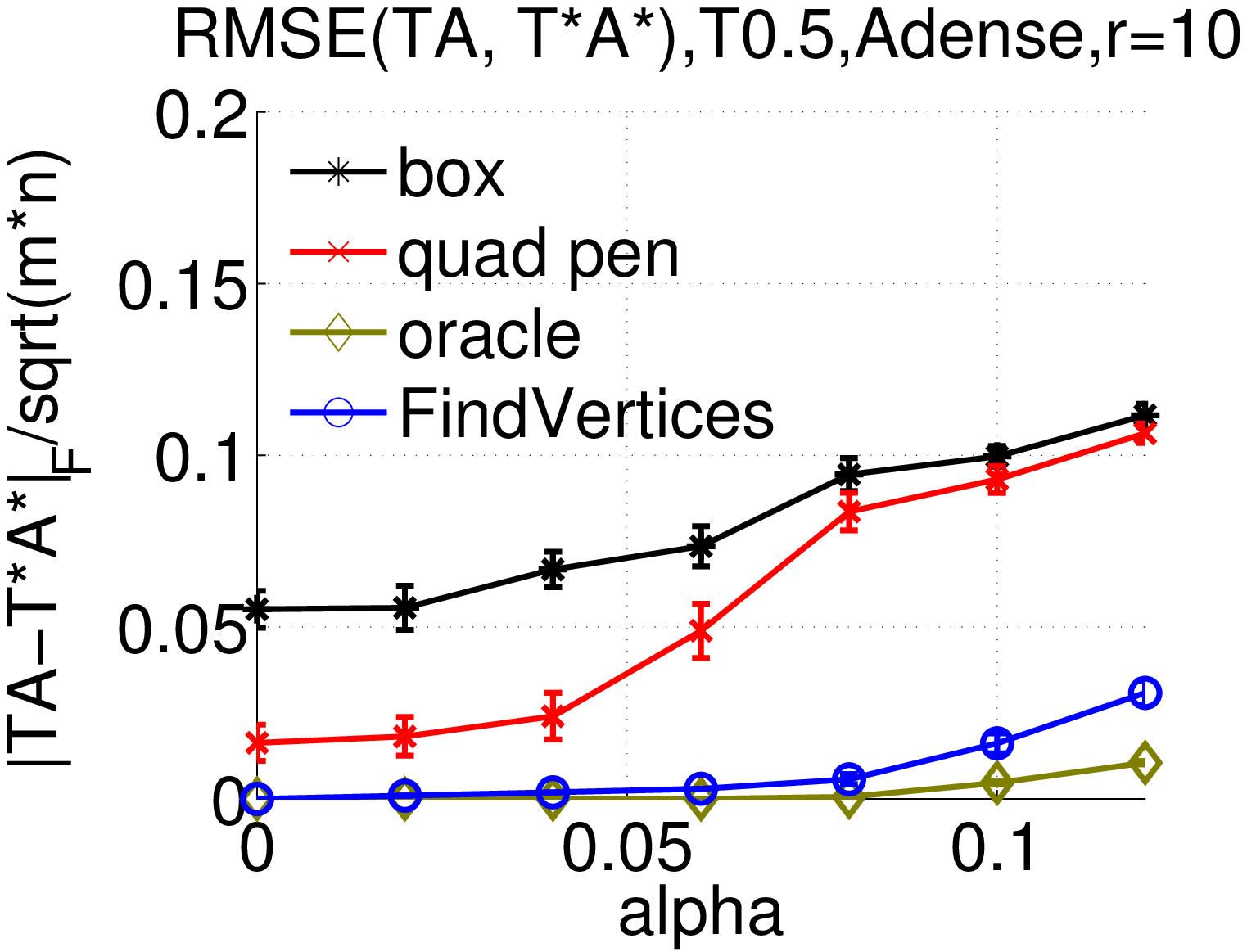}
 &\hspace{-0.29cm} \includegraphics[height =
 0.15\textheight]{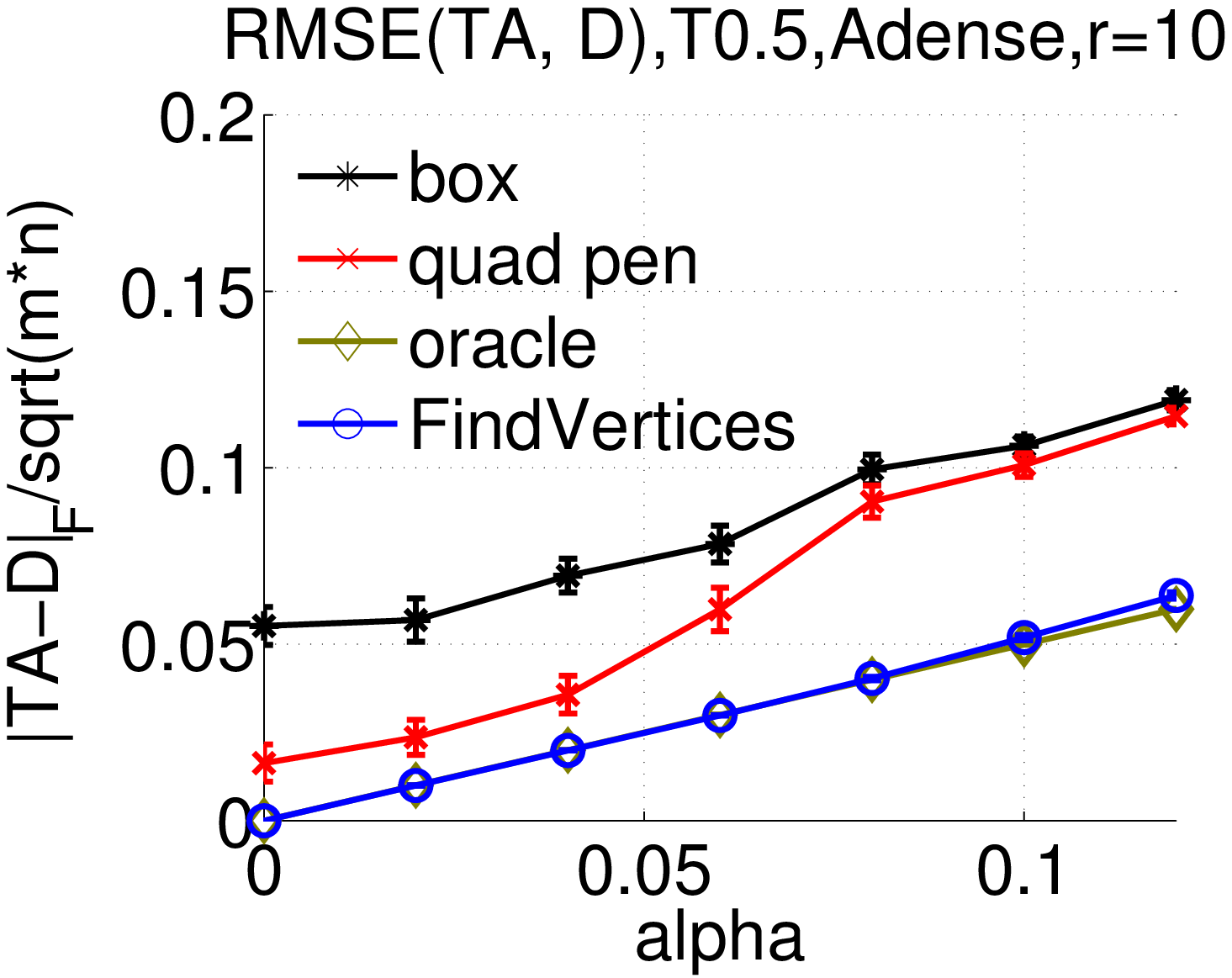}\\
 \hspace{-0.28cm}\includegraphics[height = 0.15\textheight]{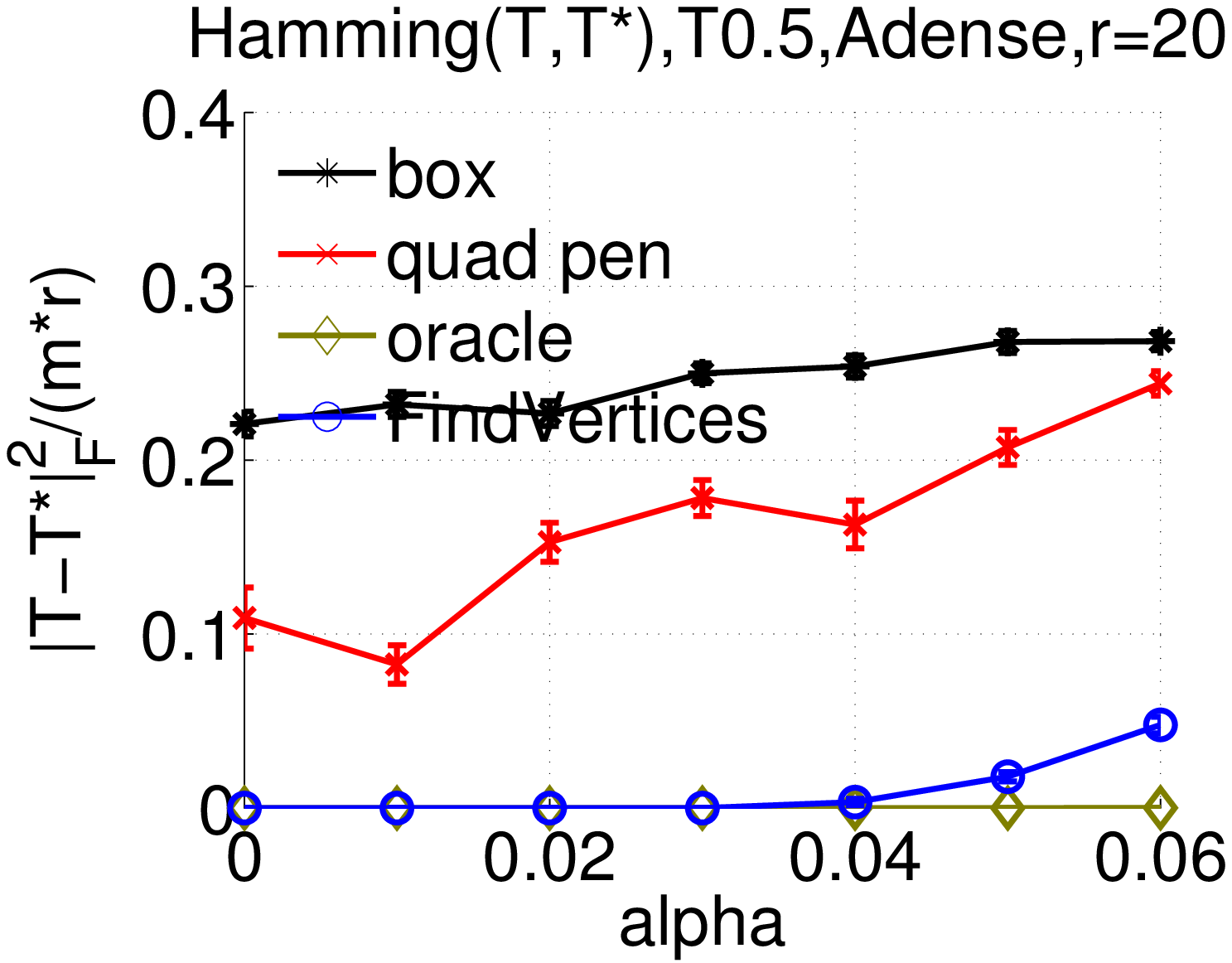}
 & \hspace{-0.28cm}\includegraphics[height = 0.15\textheight]{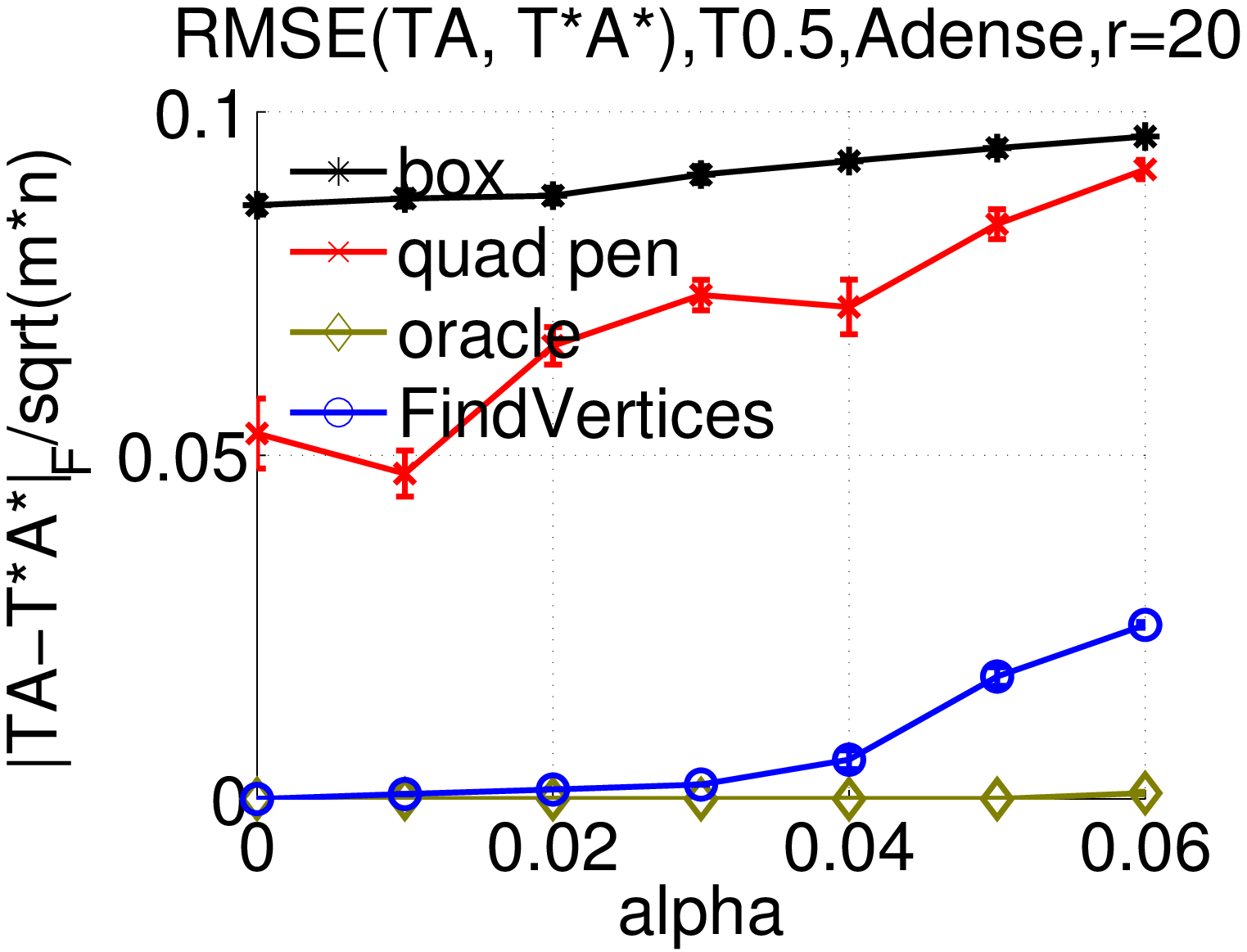}
 & \hspace{-0.29cm}\includegraphics[height =
 0.15\textheight]{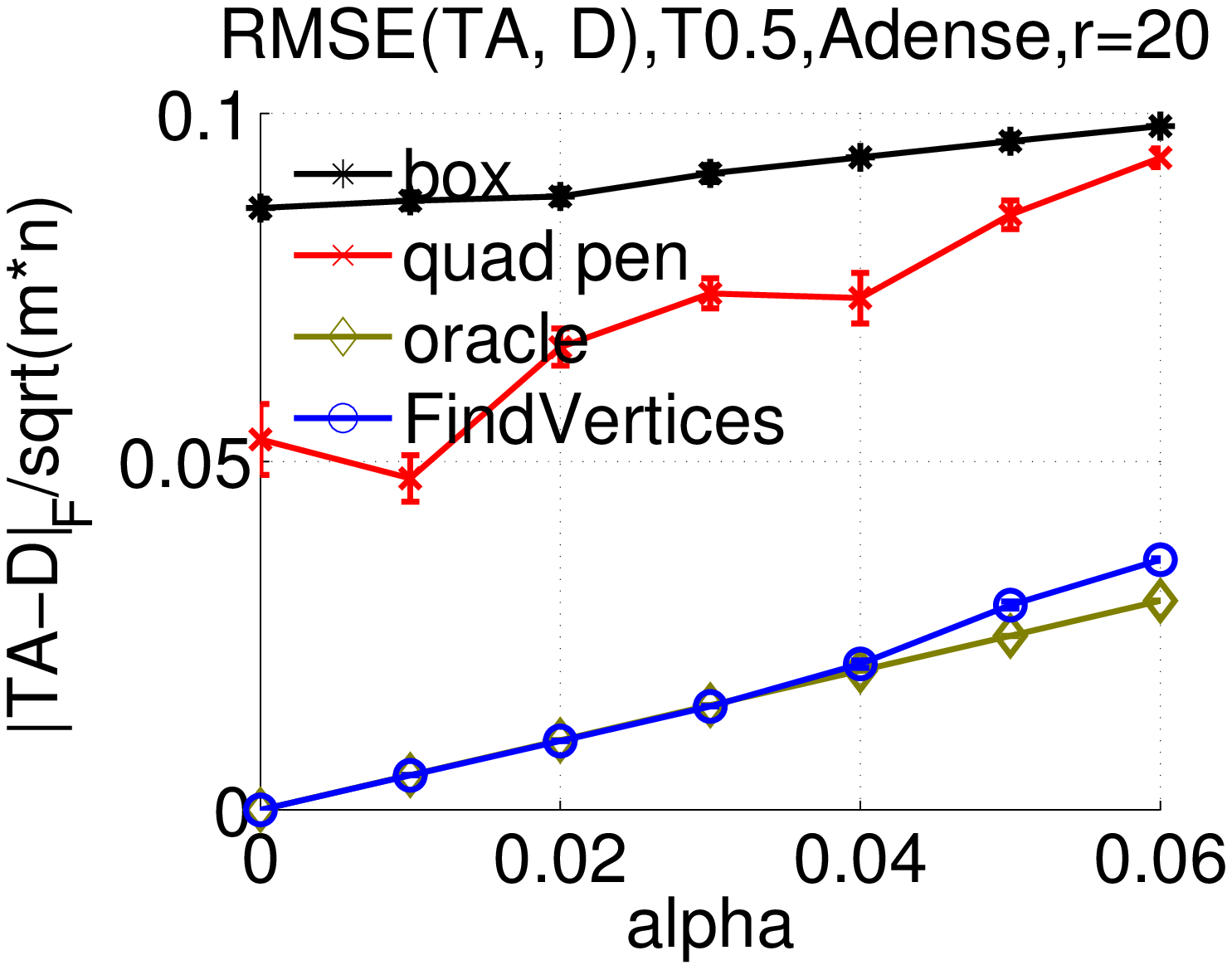}\
\end{tabular}
\caption{Results of the synthetic data experiments separated according to the
  setups \emph{'T.05'}, \emph{'Tsparse+dense'} and \emph{'T0.5,Adense'}. Bottom/top:
  $r=10$, $r=20$. Left/Middle/Right: $\nnorm{T^* - T}_F^2 / (m\,r)$, $\nnorm{T^*A^* - TA}_F/(m \,
n)^{1/2}$ and $\nnorm{T A - D}_F/(m \, n)^{1/2}$.}\label{fig:toy1}
\end{flushleft}
\end{figure} 
\clearpage 
Regarding the comparison against \textsf{HOTTOPIXX}, only the results for $r = 10$
are reported in the paper. We here display the results for $r = 20$ as well.\hfill\\
\begin{figure}[h!]
\begin{flushleft}
\begin{tabular}{lll}
\hspace{-0.28cm}\includegraphics[height = 0.15\textheight]{errT_100_k10.eps}
&\hspace{-0.28cm} \includegraphics[height = 0.15\textheight]{denoising_100_k10.eps}
&\hspace{-0.29cm} \includegraphics[height =
0.15\textheight]{objective_100_k10.eps}\\
\hspace{-0.28cm}\includegraphics[height = 0.15\textheight]{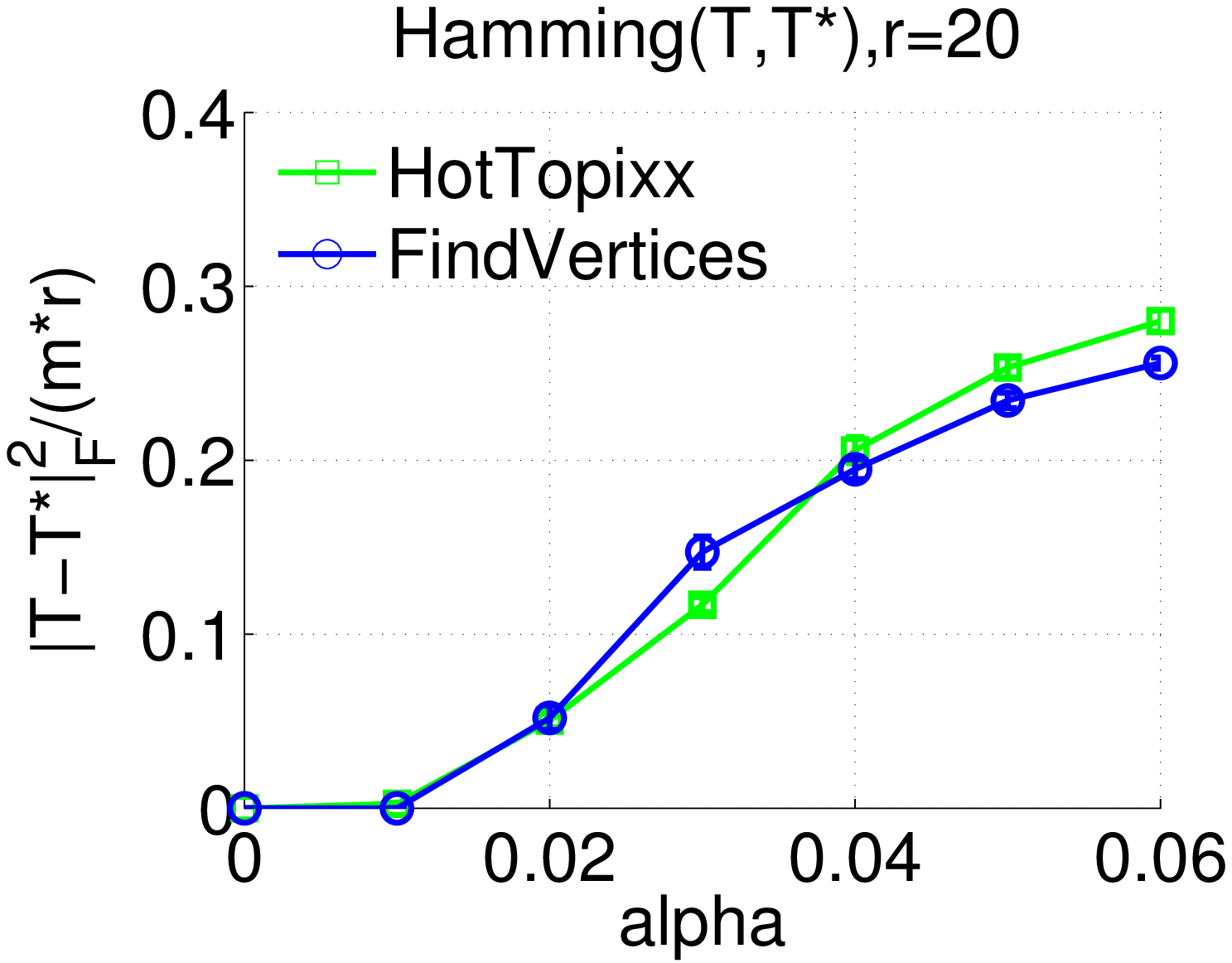}
& \hspace{-0.28cm}\includegraphics[height = 0.15\textheight]{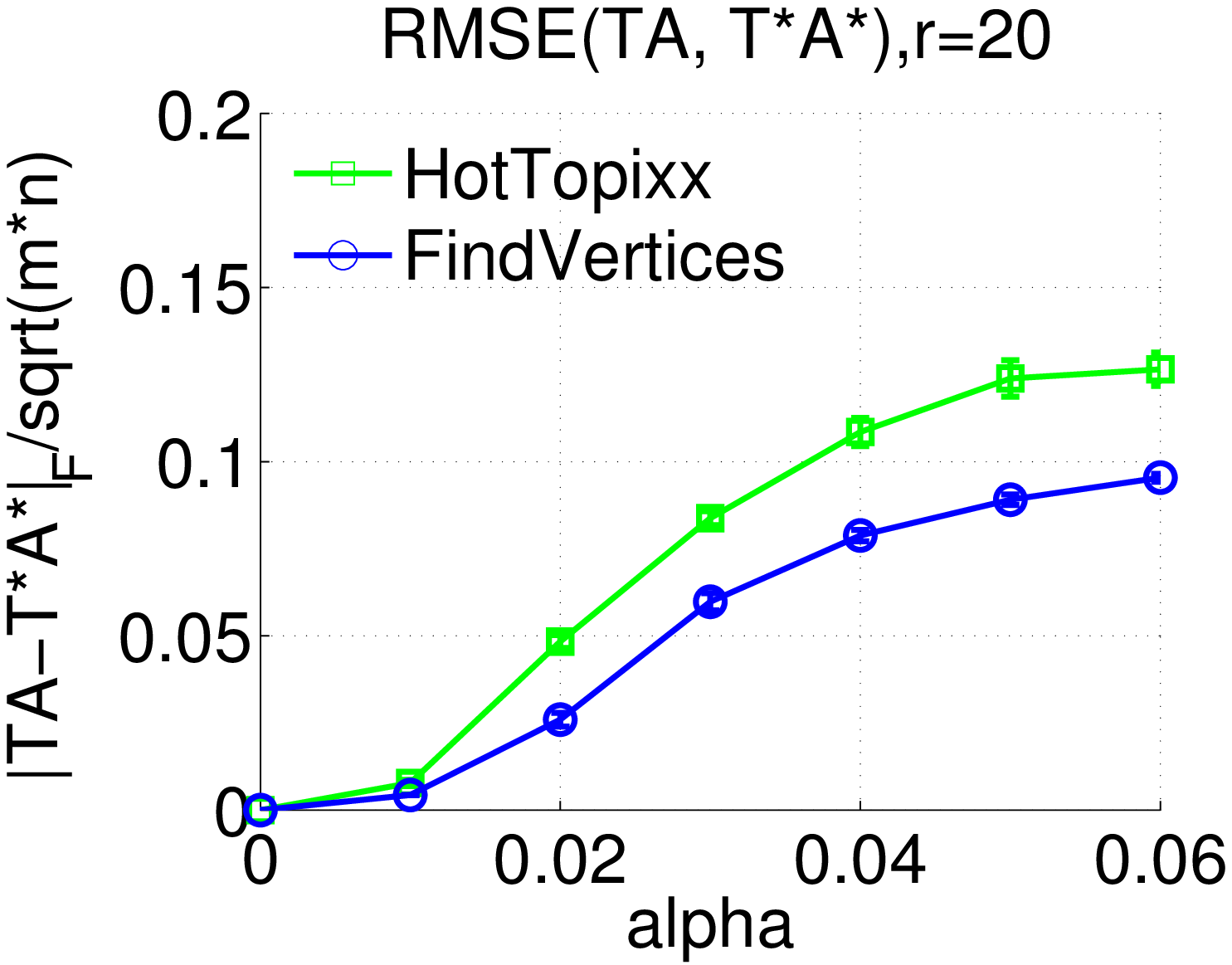}
& \hspace{-0.29cm}\includegraphics[height =
0.15\textheight]{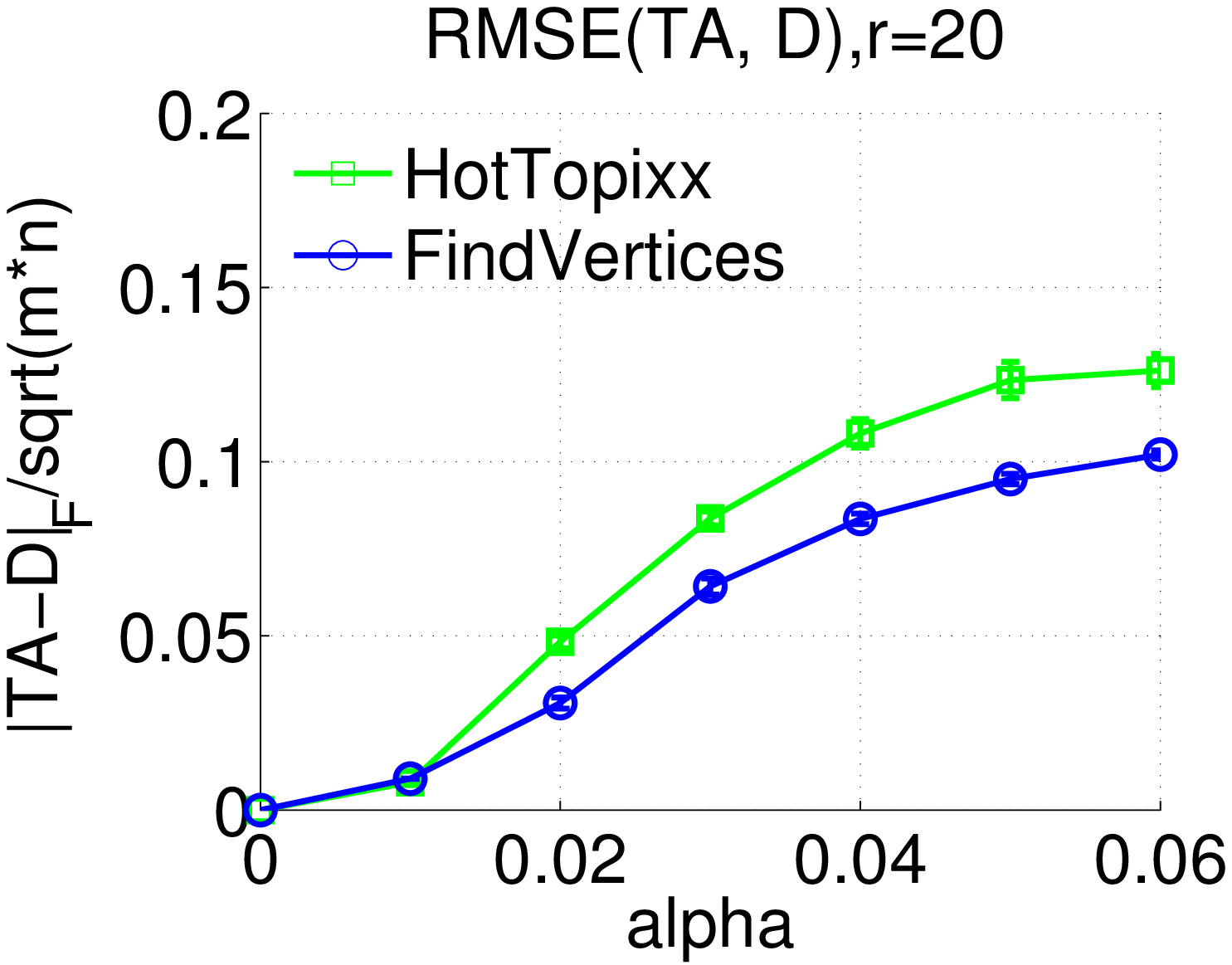}\\
\end{tabular}
\caption{Results of the experimental comparison against \textsf{HOTTOPIXX}.}\label{fig:hottopixx}
\end{flushleft}
\end{figure}

%













\clearpage

{\footnotesize
\bibliographystyle{unsrt}

\bibliography{/warehouse/Proteomics/qinqing/hypercube/paper_rev/ref_regfac}
}

\end{document}